\documentclass[twocolumn, switch]{article} 

\usepackage{preprint}

\usepackage{amsmath, amsthm, amssymb, amsfonts}

\usepackage[authoryear,round]{natbib}
\usepackage[utf8]{inputenc}	
\usepackage[T1]{fontenc}	
\usepackage{xcolor}		
\usepackage{hyperref}	
\usepackage{booktabs} 		
\usepackage{nicefrac}		
\usepackage{microtype}		
\usepackage{lineno}		
\usepackage{float}			

\usepackage{newfloat}
\DeclareFloatingEnvironment[name={Supplementary Figure}]{suppfigure}
\usepackage{sidecap}
\sidecaptionvpos{figure}{c}

\usepackage{titlesec}
\titlespacing\section{0pt}{12pt plus 3pt minus 3pt}{1pt plus 1pt minus 1pt}
\titlespacing\subsection{0pt}{10pt plus 3pt minus 3pt}{1pt plus 1pt minus 1pt}
\titlespacing\subsubsection{0pt}{8pt plus 3pt minus 3pt}{1pt plus 1pt minus 1pt}

\usepackage{amsmath}
\usepackage{graphicx}

\usepackage{color}
\usepackage{xcolor}
\usepackage{comment}
\usepackage{gensymb}
\usepackage{svg}
\usepackage{booktabs}
\usepackage{tabularx}
\usepackage{enumitem}
\usepackage[labelfont=bf]{caption}

\setlist[itemize]{leftmargin=1.5em}
\DeclareMathOperator*{\concat}{\scalebox{1}[1.5]{$\parallel$}}

\usepackage{tikz,xcolor,hyperref}

\definecolor{lime}{HTML}{A6CE39}
\DeclareRobustCommand{\orcidicon}{
	\begin{tikzpicture}
	\draw[lime, fill=lime] (0,0) 
	circle [radius=0.16] 
	node[white] {{\fontfamily{qag}\selectfont \tiny ID}};
	\draw[white, fill=white] (-0.0625,0.095) 
	circle [radius=0.007];
	\end{tikzpicture}
	\hspace{-2mm}
}
\foreach \x in {A, ..., Z}{\expandafter\xdef\csname orcid\x\endcsname{\noexpand\href{https://orcid.org/\csname orcidauthor\x\endcsname}
			{\noexpand\orcidicon}}
}

\title{U-GAT: Multimodal Graph Attention Network for COVID-19 Outcome Prediction}

\usepackage{xwatermark}
\newwatermark[firstpage,color=black!70,angle=0,scale=0.28, xpos=0in,ypos=-5in]{*Corresponding author: \texttt{\href{mailto:matthias.keicher@tum.de}{\color{black}{matthias.keicher@tum.de}}}, \\$^1$ Contributed equally.}

\usepackage{authblk}

\author[a]{Matthias Keicher\textsuperscript{*,$1$,}}
\author[a]{Hendrik Burwinkel\textsuperscript{$1$,}}
\author[a]{David Bani-Harouni\textsuperscript{$1$,}}
\author[a]{Magdalini Paschali}
\author[a]{Tobias Czempiel}
\author[b,c]{Egon Burian}
\author[c]{Marcus R. Makowski}
\author[c]{Rickmer Braren}
\author[a,d]{Nassir Navab}
\author[a]{Thomas Wendler}

\affil[a]{Computer Aided Medical Procedures, Technical University of Munich}
\affil[b]{Department of Diagnostic and Interventional Radiology, School of Medicine, Technical University of Munich}
\affil[c]{Department of Diagnostic and Interventional Neuroradiology, School of Medicine, Technical University of Munich}
\affil[d]{Computer Aided Medical Procedures, Johns Hopkins University}

\begin{document}

\twocolumn[ 
  \begin{@twocolumnfalse} 
  
\maketitle

\begin{abstract}
During the first wave of COVID-19, hospitals were overwhelmed with the high number of admitted patients. An accurate prediction of the most likely individual disease progression can improve the planning of limited resources and finding the optimal treatment for patients. However, when dealing with a newly emerging disease such as COVID-19, the impact of patient- and disease-specific factors (e.g. body weight or known co-morbidities) on the immediate course of disease is by and large unknown. In the case of COVID-19, the need for intensive care unit (ICU) admission of pneumonia patients is often determined only by acute indicators such as vital signs (e.g. breathing rate, blood oxygen levels), whereas statistical analysis and decision support systems that integrate all of the available data could enable an earlier prognosis. To this end, we propose a holistic graph-based approach combining both imaging and non-imaging information. Specifically, we introduce a multimodal similarity metric to build a population graph for clustering patients and an image-based end-to-end Graph Attention Network to process this graph and predict the COVID-19 patient outcomes: admission to ICU, need for ventilation and mortality. Additionally, the network segments chest CT images as an auxiliary task and extracts image features and radiomics for feature fusion with the available metadata. Results on a dataset collected in Klinikum rechts der Isar in Munich, Germany show that our approach outperforms single modality and non-graph baselines. Moreover, our clustering and graph attention allow for increased understanding of the patient relationships within the population graph and provide insight into the network's decision-making process.
\end{abstract}

\vspace{0.35cm}

  \end{@twocolumnfalse} 
] 



\begin{figure*}[h]
    \centering
    \includegraphics[width=\textwidth]{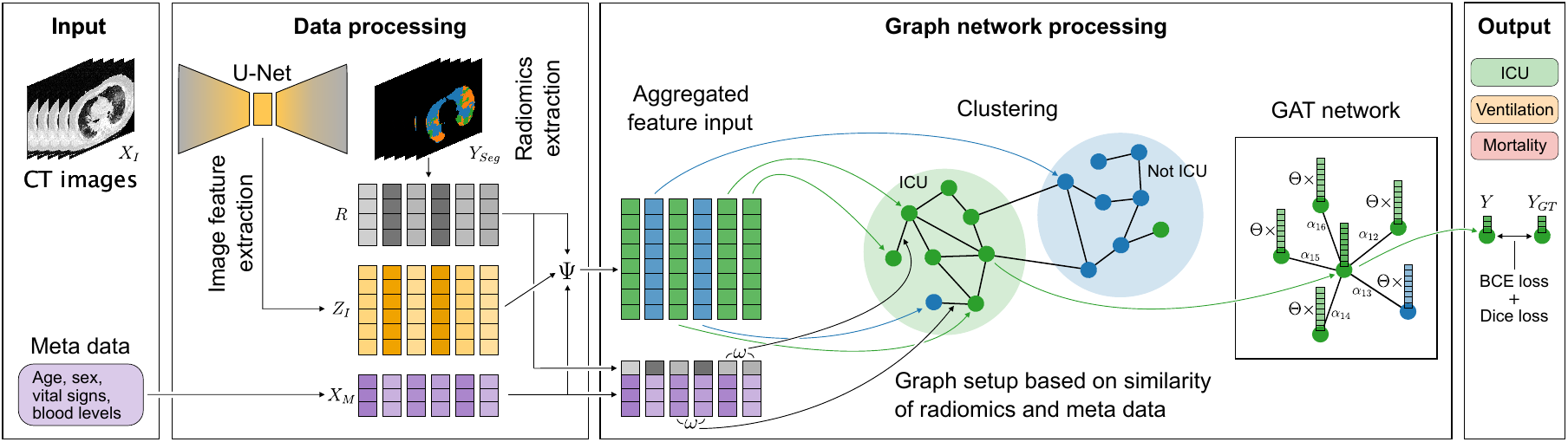}
    \caption{U-GAT is an end-to-end trained model that combines image and radiomic features ($Z_I$ and $R$ respectively) with clinical metadata $X_M$ like age, sex, vital signs and blood levels to predict the patient outcome of a disease. The segmentation $Y_{Seg}$ of disease-affected areas in a CT image $X_I$ is used as an auxiliary task to extract the radiomic features $R$ from the segmentation output and to regularize the extraction of the image feature $Z_I$ from the bottleneck of the U-Net. All features are concatenated using a function $\Psi$ into a multimodal feature vector. Test patients are clustered in a graph with training patients using the distance $\omega$ of radiomic and metadata features. Finally, a Graph Attention Network (GAT) refines the features to predict the most likely patient outcome $Y$ and using the learned linear transformation $\Theta$, as well as the attention coefficients $\alpha_{ij}$ between patients. We use binary cross-entropy (BCE) to compare $Y$ with the outcome ground truth $Y_{GT}$. For the segmentation auxiliary task, we use a Dice loss with manually annotated ground truth. Applied for COVID-19, we segment pathologies in lung CT images and predict the ICU admission, need for ventilation and survival of patients. The graph construction is only depicted for the binary ICU prediction, but can be extended to a multilabel setting.}
    \label{fig:unet_scheme}
\end{figure*}

\section{Introduction}
During the first wave of the coronavirus disease 2019 (COVID-19)~\citep{wang2020novel}, the exponential growth in new cases led to fast overcrowding of intensive care units (ICU) resulting in never before seen scenes in the age of modern medicine~\citep{remuzzi_COVID-19_2020,ryberg_covid-19_2021}. 
During such a state of emergency, an optimization of the allocation of hospital resources, e.g., ICU beds, mechanical ventilators or personnel becomes crucial. An essential aspect of effective patient management is a correct assessment of treatment necessity and potential outcome. Due to the limited understanding of the previously unknown disease paired with a highly multimodal data situation, performing such an assessment and prediction of patient outcome is however very challenging. The resulting sudden overload of care facilities in combination with the high complexity of the obtained data structure motivates the need for assistance systems for fast outcome prediction and triaging based on available patient information.\\
Upon a patient's hospital admittance, a multitude of parameters - such as sex, age, body weight, symptoms, co-morbidities, blood cell counts, inflammatory parameters, biochemical values, cytokine profiles among others - are obtained and documented. These parameters - referred to as "metadata" in the following - as well as, radiological images including radiographs or X-ray computed tomography (CT) images, are available within the first hours after a new patient arrives at the hospital. This deems them ideal for an early triaging and outcome prediction. Classical disease outcome prognosis is based, however, also on anamnestic information and clinical experiences. The general logic of a physician's decision-making process partly relies on the information embedded in similar patients where the outcome and the connection of this information to currently treated patients are known. Such population relationships are particularly useful when little to no disease-specific epidemiological information as well as deep medical expertise is available, as is the case in a novel pandemic.\\
Following this method of reasoning, we propose a decision support system that performs a multimodal data analysis to cluster COVID-19 patients in a population graph which is then used in a graph neural network with an attention mechanism to refine the patient outcome prediction by taking into account similar patients. The used similarity metric and attention mechanism provide added insight into the decision-making process. This becomes possible as the weighting of clinical features and the most influential patients used in the prediction process can be directly observed. Our contributions are as follows:
\begin{itemize}
    \item We introduce U-GAT, an end-to-end graph-based method for predicting COVID-19 patient outcomes, namely ICU admission, need for ventilation and mortality. Our model uses a U-Net~\citep{Ronneberger2015} to segment the healthy and pathological regions of the lung in chest CTs and subsequently performs a joint feature fusion of image, radiomic and metadata features. This combined feature vector is refined in our Graph Attention Network (GAT) \citep{Velickovic2018} by leveraging similar patients to perform the final outcome prediction.
    \item We present an interpretable, multimodal similarity metric for graph construction and effective batch selection.
    \item We thoroughly evaluate our novel approach on a newly acquired dataset collected in Klinikum rechts der Isar in Munich during the first COVID-19 wave of 2020 and showcase our model's ability to predict patient-specific disease outcomes.
\end{itemize}
This article is structured in eight sections: in Section \ref{sec:related_work}, we discuss related work regarding Graph Convolutional Networks (GCNs) and methods for COVID-19 detection and outcome prediction. In Section \ref{sec:material}, the dataset collection, annotation and processing pipeline are presented. In Section \ref{sec:methodology}, we introduce our novel methodology. In Section \ref{sec:experiments}, we present the experimental setup and baselines used for our evaluation scheme. Section \ref{sec:results} presents our evaluation results for baselines and ablative studies, which we further discuss in Section \ref{sec:discussion}. Finally, Section \ref{sec:conclusion} contains the conclusions and an outlook. Theoretical fundamentals for GCNs motivating our design choices are included in Appendix~\ref{app:theory}.
\section{Related Work}
\label{sec:related_work}
\subsection{Graph Convolutional Networks for medical imaging}
In previous work Graph Convolutional Networks (GCNs) demonstrated their potential for medical applications due to their optimized processing of medical image information. One of the first works applying GCNs to a medical task~\citep{parisot2017spectral} used GCNs for improved prediction of Alzheimer's disease and Autism Spectrum Disorder (ASD).
In a follow-up work \cite{parisot2018disease} showed that depending on the included patient information for the graph setup the network performance can vary substantially, motivating a more thorough analysis of graph components within a GCN. To gain more independence from the graph setup, \cite{anirudh2019bootstrapping} proposed a bootstrapping strategy and ensemble learning for GCNs to reduce performance dependencies from the graph generation. Instead of training on one population graph, they created several graphs by randomly dropping edges from the initially designed setup.
\cite{cosmo2020latent} introduced a method for self-learned graph construction used for an optimized learning behavior of inductive GCNs, combining both imaging and non-imaging data.\\
The aforementioned works leveraged already extracted image features, however, \cite{Burwinkel2019} proposed a methodology that used GCNs on image data directly. They showed that end-to-end processing of imaging and metadata within a GCN can lead to improved performance due to optimized feature learning. At the same time, the proposed approach allowed for more effective usage of inter-class connections within the graph. We will expand upon this concept within our developed methodology as well and explain the described implications in detail in Sec.~\ref{sec:methodology}.
Also focusing on improved image classification, \cite{du2019zoom} proposed a two-step approach to perform mammogram analysis, leveraging GATs. First, they used a node classification to focus on a region of interest (ROI) and then a graph classification to determine whether the ROI was malignant or benign. GCNs have also been used for medical segmentation of the pancreas \citep{soberanis2020uncertainty}, prostate \citep{tian2020graph}, fundus and ultrasound images \citep{meng2020cnn} as well as cardiac CT angiography \citep{wolterink2019graph}.

\subsection{GCNs for COVID-19}
In the context of COVID-19 diagnosis, GCNs have mainly been adapted for disease detection. \cite{wang2021covid} and \cite{yu2020resgnet} built graphs based on the similarity of extracted CT image features and classified the nodes for the presence of infiltrates. In addition to image features, \cite{song2021augmented} and \cite{liang2021diagnosis} used the site of acquisition along with other features to build a multi-center graph to combat domain shift and improve COVID-19 detection. Instead of modeling a patient population, \cite{saha2021graphcovidnet} converted edges detected in chest CT and X-ray images to graphs and leveraged these for detecting COVID-19. \cite{huang2021graph} used GCNs to refine the bottleneck features for the binary segmentation of COVID-19 infections. Finally, \cite{di2021hypergraph} learn an uncertainty-vertex hypergraph to distinguish between community-acquired pneumonia and COVID-19. They use regional image features and radiomics to model relationships between patients and show an improved classification accuracy over non-graph methods.
To the best of our knowledge, we propose the first graph-based end-to-end COVID-19 patient outcome prediction method by leveraging a population graph using chest CTs and patient metadata.

\subsection{Multitask learning for COVID-19}
Recent works~\citep{Burian2020, Colombi2020, shiri2021machine,wang2021study} on the radiological assessment of COVID-19 patients have shown a high correlation between disease burden and patient outcome, e.g. the probability of ICU admission. Several deep learning methods have been proposed that try to exploit this correlation with multitasking approaches~\citep{yang2017novel, mehta2018net,le2019multitask}. 
The majority of the proposed multitask methods focus on the joint detection of COVID-19 infection and the binary segmentation of related pathologies in lung CT images.
~\cite{Alom2020, Wu2020} deployed separate models for the detection of COVID-19 and the segmentation of the diseased area in the lung.
~\cite{Amyar2020} used a single U-Net-based multitask framework consisting of three heads for the segmentation, classification and reconstruction of the input image. \cite{Gao2021} deployed a dual-branch network utilizing a U-Net~\citep{Ronneberger2015} branch for segmentation and a ResNet50~\citep{He2016} branch for classification, that interacted with an attention mechanism.\\
Towards COVID-19 patient outcome prediction, another set of works applied multitasking on the joint estimation of the severity of COVID-19 and various classification and segmentation tasks. \cite{He2020} performed a joint segmentation of the lung lobes and a binary prediction of the severity of COVID-19 using a U-Net-based multiple-instance learning method.
\cite{goncharov2021ct} combined the segmentation of the lung and COVID-19 pathologies with the identification of COVID-19 cases and the estimated severity. 
In contrast to most other multitask methods, they branched the final feature maps of a U-Net architecture for the two classification tasks instead of the low-resolution bottleneck representation that most approaches used for classification, reporting an improved performance. \cite{Bao2020} introduced a multitask framework built on a SqueezeNet~\citep{SqueezeNet} backbone for the segmentation of the lung and two classification tasks: the detection of COVID-19 and the estimation of three degrees of severity.
Similar to our approach \cite{nappi2021u} use the bottleneck features of a pretrained U-Net for the prediction of COVID-19 progression and mortality. However, they did not utilize a graph-based approach for the classification.

\subsection{Fusing imaging and metadata for COVID-19}
\label{sec:intro_feature_fusion}
Motivated by the decision-making process performed by an experienced physician and medical boards that combine domain-specific knowledge from different fields of expertise, the development of machine learning systems that operate in a comparable manner finds growing interest in research. \cite{Huang2020} defined three types of methods for the fusion of features in deep learning models for radiological applications: fusion of extracted image features with non-imaging features (early fusion), feature fusion with a joint end-to-end (image) feature extraction (joint fusion) and the aggregation of predictions made by independent models (late fusion). \\
For the COVID-19 detection and the prediction of patient outcome, most of the proposed methods integrating both imaging and non-imaging data apply early fusion of features: \cite{Burian2020} and \cite{Chao2021} combine radiomics extracted using a semi-automatic lung CT segmentation with clinical data in a Random Forest (RF) for the prediction of ICU admission. Using similar methods and data, \cite{Tang2021} distinguished between severe and non-severe cases and ~\cite{Cai2020} predicted different degrees of recovery and survival rates of patients. Processing tabular data only, \cite{Xu2020} evaluated the performance of different machine learning methods for the prediction of ICU admission, need for ventilation and survival using radiomics, clinical and laboratory data with the best performance using an aggregation of all available data modalities. \cite{Homayounieh2020} showed an improvement in the prediction of COVID-19 patient outcome (recovery or death)
by the addition of clinical variables to extracted radiomics features in a Random Forest model. 
\cite{Chassagnon2021} demonstrated the importance of combining a wide range of non-imaging and extracted imaging features for the short- and long-term prognosis of COVID-19 patients in an ensemble of multiple classical machine learning models. \cite{shiri2021machine} performed uni- and multivariate analysis of clinical data and lung CT radiomics and classified the survival of COVID-19 patients with the gradient boosting library XGBoost \citep{chen2016xgboost}. Their best results were achieved by combining lesion-specific radiomics and clinical data. \cite{gong2021multi} improved the results of their generalized linear model for predicting severe COVID-19 outcomes by adding blood values to other clinical features and extracted radiomics.\\
Applying late fusion with penalized logistic regression, \cite{Ning2020} reported an improvement of both COVID-19 severity and mortality outcome prediction compared to the stand-alone lung CT Convolutional Neural Network (CNN) and non-imaging Multilayer Perceptron (MLP) models.
\cite{tariq2021patient} explored different fusion methods for predicting the need for hospitalization of COVID-19 patients and found the early fusion of different electronic medical record features to work best for this task.\\
A joint fusion method combining imaging and non-imaging data to predict ICU admission, ventilation and mortality has not been proposed yet.

\begin{table*}[!htp]
    \caption{\label{tab:ICU_blood}Blood values at admission for the 53 patients who were admitted to the ICU and for the 79 patients who were not. The sum of $n$ differs from our total amount of patients ($n=132$) due to missing values for some patients. * denotes the significant difference assuming a $5\%$ significance level.}
  \centering
  \vspace{3pt}
  \begin{tabular}{llllllll}
    \toprule
    & \multicolumn{3}{c}{ICU (n=53)} & \multicolumn{3}{c}{No ICU (n=79)} & \\
    \cmidrule(lr){2-4} \cmidrule(lr){5-7}
    Blood value & Average & Std. Dev. & n & Average & Std. Dev. & n & p \\
    \midrule
Leukocytes (G/L)	&8.4	&4.9	&53	&6.7	&4.1	&79	&0.03*\\
Lymphocytes (G/L)	&19.3	&46.9	&48	&22.6	&35.3	&75	&0.65\\
Thrombocytes (G/L)	&226.6	&100.1	&53	&228.5	&116.8	&79	&0.92\\
C-reactive protein (CRP. mg/dL)	&12.19	&9.30	&53	&6.10	&6.26	&78	& $<$0.01*\\
Creatinine (mg/dL)	&1.56	&1.67	&53	&4.17	&26.50	&78	&0.48\\
D-Dimer ($\mu$g/mL)	&5467	&12801	&41	&1952	&5570	&67	&0.05\\
Lactate dehydrogenase (LDH. U/L)	&468.6	&329.5	&48	&358.4	&368.4	&75	&0.09\\
Creatinine kinase (U/L)	&427.3	&1167.2	&48	&225.3	&777.2	&74	&0.25\\
Troponine-T (ng/mL)	&0.071	&0.161	&25	&0.097	&0.323	&34	&0.71\\
Interleukin 6 (IL-6. pg/mL)	&120.5	&117.5	&35	&104.1	&388.7	&60	&0.81\\
    \bottomrule
  \end{tabular}
\end{table*}

\begin{table*}[!htp]
    \caption{\label{tab:Vent_blood}Blood values at admission for the 38 patients that needed ventilation and for the 94 that did not. The sum of $n$ differs from our total amount of patients ($n=132$) due to missing values for some patients. * denotes significant difference assuming a $5\%$ significance level.}
  \centering
  \vspace{3pt}
  \begin{tabular}{llllllll}
    \toprule
    & \multicolumn{3}{c}{Ventilation (n=38)} & \multicolumn{3}{c}{No Ventilation (n=94)} & \\
    \cmidrule(lr){2-4} \cmidrule(lr){5-7}
    Blood value & Average & Std. Dev. & n & Average & Std. Dev. & n & p \\
    \midrule
Leukocytes (G/L)	&7.9	&3.7	&38	&7.2	&4.8	&94	&0.44\\
Lymphocytes (G/L)	&13.1	&7.9	&35	&24.6	&46.8	&88	&0.15\\
Thrombocytes (G/L)	&209.8	&102.3	&38	&235.0	&112.7	&94	&0.23\\
C-reactive protein (CRP. mg/dL)	&13.54	&9.80	&38	&6.53	&6.43	&93	& $<$0.01*\\
Creatinine (mg/dL)	&1.42	&0.66	&38	&3.81	&24.28	&93	&0.55\\
D-Dimer ($\mu$g/mL)	&5622	&14484	&29	&2429	&6019	&79	&0.11\\
Lactate dehydrogenase (LDH. U/L)	&454.1	&241.9	&36	&379.6	&393.5	&87	&0.29\\
Creatinine kinase (U/L)	&533.7	&1354.8	&35	&212.7	&718.0	&87	&0.09\\
Troponine-T (ng/mL)	&0.049	&0.089	&19	&0.104	&0.316	&40	&0.47\\
Interleukin 6 (IL-6. pg/mL)	&138.2	&126.4	&24	&100.7	&358.5	&71	&0.62\\
    \bottomrule
  \end{tabular}
\end{table*}

\begin{table*}[!htp]
    \caption{\label{tab:Surv_blood}Blood values at admission for the 113 patients who survived and for the 19 who passed. The sum of $n$ differs from our total amount of patients ($n=132$) due to missing values for some patients. * denotes significant difference assuming a $5\%$ significance level.}
  \centering
  \vspace{3pt}
  \begin{tabular}{llllllll}
    \toprule
    & \multicolumn{3}{c}{Passed (n=19)} & \multicolumn{3}{c}{Survived (n=113)} & \\
    \cmidrule(lr){2-4} \cmidrule(lr){5-7}
    Blood value & Average & Std. Dev. & n & Average & Std. Dev. & n & p \\
    \midrule
Leukocytes (G/L)	
&9.8	&6.7	&19		
&7.0	&3.9	&113&0.01*\\
Lymphocytes (G/L)	&11.3	&8.3	&17	&22.9	&42.9	&106	&0.27\\
Thrombocytes (G/L)	&201.4	&99.2	&19	&232.2	&111.6	&113	&0.26\\
C-reactive protein (CRP. mg/dL) &11.06	&8.98	&19		&8.14	&7.99	&112	&0.15\\
Creatinine (mg/dL)	&1.57	&0.79	&19	&3.37	&22.13	&112	&0.72\\
D-Dimer ($\mu$g/mL)	&6388	&12076	&13	&2862	&8644	&95	&0.19\\
Lactate dehydrogenase (LDH. U/L)	&607.3	&500.1 &17	&368.4	&318.7	&106	&0.01*\\
Creatinine kinase (U/L)&843.2	&1878.8	&17		&217.6	&673.0	&105	&0.01*\\
Troponine-T (ng/mL)&0.120	&0.216	&13	&0.077	&0.279	&46		&0.61\\
Interleukin 6 (IL-6. pg/mL)&143.7	&181.9	&11		&105.8	&330.1	&84	&0.71\\
    \bottomrule
  \end{tabular}
\end{table*}

\begin{table*}[!htp]
    \caption{\label{tab:Radiomics}Radiomics derived from manually annotated at admission for patients. * = significant difference assuming a $5\%$ significance level.}
  \centering
  \vspace{3pt}
  \begin{tabular}{llllll}
    \toprule
    Radiomic & Average & Std. Dev. & Average & Std. Dev. & p \\
    \midrule \midrule
    & \multicolumn{2}{c}{ICU (n=53)} & \multicolumn{2}{c}{No ICU (n=79)} & \\
    \cmidrule(lr){2-3} \cmidrule(lr){4-5}
Healthy lung	&65.2\%	&25.9\%	&92.1\%	&9.2\%	&$<$0.01*\\
GGO	&22.7\%	&16.4\%	&6.2\%	&7.1\%	&$<$0.01*\\
Other pathologies	&12.1\%	&14.0\%	&1.9\%	&4.2\%	&$<$0.01*\\
\midrule    
& \multicolumn{2}{c}{Ventilation (n=38)} & \multicolumn{2}{c}{No Ventilation (n=94)} & \\
\cmidrule(lr){2-3} \cmidrule(lr){4-5}
Healthy lung	&61.2\%	&22.5\%	&89.4\%	&16.1\%	&$<$0.01*\\
GGO	&25.7\%	&14.9\%	&7.7\%	&10.1\%	&$<$0.01*\\
Other pathologies	&13.1\%	&13.1\%	&3.2\%	&7.9\%	&$<$0.01*\\
    \midrule
& \multicolumn{2}{c}{Passed (n=19)} & \multicolumn{2}{c}{Survived (n=113)} & \\
\cmidrule(lr){2-3} \cmidrule(lr){4-5}
Healthy lung &70.0\%	&20.5\%	&83.2\%	&22.0\%		&0.02*\\
GGO		&22.0\%	&17.2\%	&11.1\%	&13.2\%	&$<$0.01*\\
Other pathologies	&8.0\%	&9.7\%		&5.7\%	&10.8\%	&0.39\\
    \bottomrule
  \end{tabular}
\end{table*}
\section{Material and data preparation}
\label{sec:material}
\subsection{Dataset}
All experiments were conducted using a 132-patient in-house dataset, expanding on the dataset with 65 patients described in \cite{Burian2020}. The data was collected retrospectively with the consent of the local institutional review board (ethics approval 111/20 S-KH). All 132 patients (aged 24-99 years, avg. 63 years; 88 males, 44 females) were hospitalized at our institution between April 3rd and September 5th, 2020. For all these patients COVID-19 was confirmed by the polymerase chain reaction (PCR) test. 53 patients had to be admitted to the ICU for further treatment, of these 38 required machine-assisted ventilation while 19 deceased.\\
At admission patients presented fever in $66\%$, coughing in $45\%$, shortness of breath in $33\%$ and gastrointestinal symptoms in $15\%$ of the cases, respectively. Percutaneous oxygen saturation was $93.4\pm7.1\%$ and the temperature was $37.7\pm1.0\degree C$. Oxygen saturation was significantly different in ICU and non-ICU patients ($90.7\pm10.2\%$ vs. $95.0\pm3.5\%$), as well as between patients requiring ventilation and not ($89.4\pm10.5\%$ vs. $94.8\pm4.9\%$). Blood tests were made on admission and can be seen in Tables~\ref{tab:ICU_blood}, \ref{tab:Vent_blood} and \ref{tab:Surv_blood} with statistics and t-test results.\\
The non-contrast low-dose lung CT images were acquired using a 256-row multidetector computed tomography (MDCT) scanner (iCT, Philips Healthcare, Best, The Netherlands) in full inspiration with arms elevated.
To assess the patient outcome different parameters were collected: the admission to the ICU, the necessity of mechanical ventilation and the survival of the patient.\\
The complete dataset is available on request for research purposes in the frame of the BFS project AZ-1429-20C.

\subsection{Annotations}
For each CT volume, the total lung, healthy lung tissue, ground-glass opacifications (GGO), consolidations and pleural effusions area were annotated by expert radiologists (4-8 years experience). The unequivocal voxel-wise distinction of pleural effusion from consolidation was considered a difficult task even for senior reviewers. Combined with the observation that the pleural effusion class was only present in 38 of the 132 patients and if present only accounted in average for $4.1\%$ of the lung volume ($1.2\%$ for all patients) we decided to combine the pleural effusion and consolidation class into a single class named \textit{other pathologies}. 

\subsection{Data preprocessing}
The CT images were provided with varying sub-millimeter spacing and resampled to an isotropic spacing of $3.6$ mm, resulting in an image size of $96\times96\times96$ voxels. We clipped the Hounsfield in a range of $-1024$ and $150$ for improved contrast of the lung cavity and further applied a patient-level min-max normalization by subtracting the minimum value and dividing by the maximum value of the volume.
All numerical input data such as clinical metadata was z-score normalized by subtracting the mean and dividing by the standard deviation. Missing values were replaced with mean imputation.

\subsection{Equidistant subsampling}
\label{sec:equidistant}
One of the core challenges when working with GCNs and end-to-end feature extraction from 3D images is the limited batch size due to high memory requirements. For the GCN to work effectively, multiple patient instances have to be present in a single batch. However, due to the large memory consumption of the parallel feature extraction of multiple 3D volumes, loading full volumes of multiple patients is not possible. Since COVID-19 can manifest in any part of the lung \citep{ye2020chest}, training on single slices or small patches of the patient volume is also not feasible. To this end, we propose an equidistant subsampling of $S$ slices per volume along the main imaging axis (axial view) for training. Given an image size of $Z$ along the main axis, each volume is split into $\lfloor Z/S \rfloor$ stacks of $S$ slices with $(Z\mod S)/2$ slices being cut off on each side. In addition to increasing the chance of sampling disease-affected areas, this method has the benefit of reducing overfitting by splitting the limited amount of 3D volumes in multiple samples per patient. At test time, a stack of all available slices is used covering the whole 3D volume.

\section{Methods}
\label{sec:methodology}
\subsection{Overview of the proposed methodology}
Our proposed method provides an effective way to process multimodal patient information such as CT images $X_I$ combined with metadata $X_M$ for outcome prediction of patients. For a COVID-19 patient admitted to the hospital, the three outcomes we predict are the need for ICU admission $Y_{\text{ICU}}$, the need of mechanical ventilation $Y_{\text{Vent}}$ and the survival of the patient $Y_{\text{Mort}}$. Additionally, we use the segmentation ground truth $Y_{\text{Seg}}$ of the CT-images $X_I$ as an auxiliary target to improve the training. We are optimizing the objective function $\Phi(X_I,X_M): (X_I,X_M) \rightarrow (Y_{\text{ICU}}, Y_{\text{Vent}}, Y_{\text{Mort}}, Y_{\text{Seg}})$, where $\Phi$ is our proposed network. From the segmentation output $Y_{\text{Seg}}$ we calculate radiomic features $R$ that represent the relative burden of each of the pathology classes in relation to the total lung volume and the relative healthy-appearing lung. To effectively incorporate the different types of patient data, we introduce a new framework that combines the segmentation capabilities of U-Net with the analytic strengths of Graph Convolutional Networks. This network makes use of a graph structure based on the patient similarity of the metadata $X_M$ and the extracted radiomic features $R$ and allows us to effectively process the provided CT-image data within this structure.
The proposed method operates end-to-end to perform an ideal combination of the image feature representation learning, U-Net image segmentation and graph data processing.

\subsection{Graph-based image processing}
To allow for inference on unseen data samples, we employ spatial graph convolutional approaches for our proposed network. Compared to spectral methods, this approach does not rely on the eigenbasis of the graph Laplacian and therefore allows an extension to unseen samples. For a detailed explanation of the underlying theory on convolutions performed on a graph structure and further motivation for this design choice, we refer the reader to Appendix~\ref{app:theory}. The main challenge is an effective combination of the different modalities. As explained in \ref{sec:intro_feature_fusion}, the image data $X_I$ is one of the most crucial factors. For GCNs, image-based information is usually first extracted either manually or with a pretrained CNN. These extracted numerical features are then, in a second step, processed within the graph network.\\
The advantage of this procedure lies in its memory efficiency since for the graph processing a larger amount of data has to be processed within the same batch for increased interaction between neighboring representation $x_{I,i} \in X_I$. Imaging data requires more memory which limits the direct processing of the images and often requires reducing the batch size, i.e. the part of the graph that can be processed. At the same time, the feature extraction process can potentially benefit from the underlying graph structure, indications for this were found in \cite{Burwinkel2019}. By performing an end-to-end feature extraction with a graph neural network, the graph information can be included in the extraction process and therefore optimize the representation for its usage within the graph. We leverage this concept for the processing of the provided CT-image information. Every CT image $x_{I,i}$ is processed by a U-Net to perform segmentation on the individual image slices. The calculated feature maps at the bottleneck of the U-Net are extracted (description in Sec.~\ref{sec:segmentation}) and processed to receive a corresponding representation $z_{I,i}$, usable within the graph neural network.

\subsection{Graph construction method}
\label{sec:exp_graph}

\begin{figure*}[t]
    \centering
    \includegraphics[width=\textwidth]{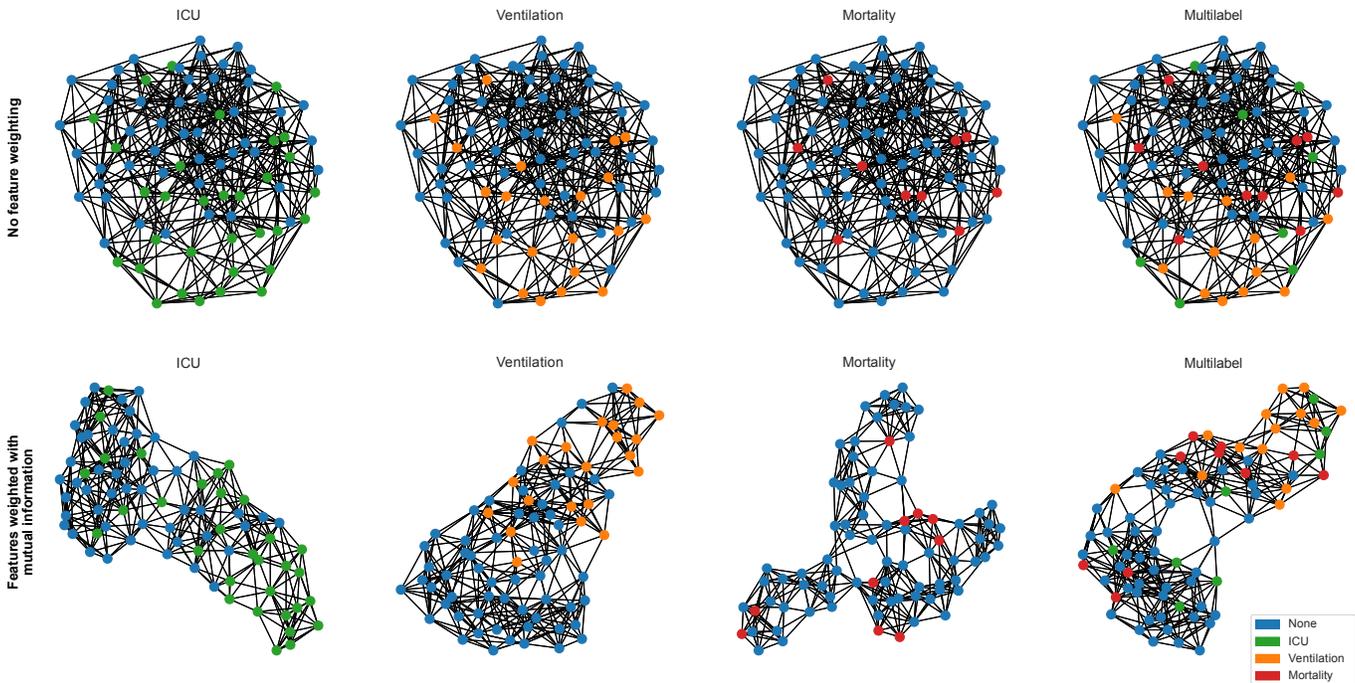}
    \caption{For an initial clustering of the patients, we use the simliarity in clinical and radiomic features. The top row shows graphs that were constructed by connecting each node with its 7 nearest neighbors using the Euclidean distance of all z-scores normalized features. The patients visualized are from the training set of a single fold. Since this results in the same graph for all tasks, the only difference are the node labels. To optimize this graph, we propose to weight each feature in the distance calculation with its mutual information with the task at hand (bottom row). This way more important features have a higher impact on the clustering and the graph can be optimized for a specific task without performing feature selection or using prior knowledge. In the multilabel setting, the mutual information is calculated with the ordinal regression of patient outcomes from less to more severe (ICU, ventilation, mortality).}
    \label{fig:graph_construction}
\end{figure*}

We define a binary, directed graph $G(V,E)$ with vertices $V$ and connecting edges $E$. Every vertex $v_i \in V$ corresponds to a stack of CT images $x_{I,i} \in X_I$ (sampling process described in Sec.~\ref{sec:equidistant}), a vector of radiomics features $r_i \in R$ (extraction process described in detail in Sec.~\ref{sec:radiomics_ex}) and patient metadata $x_{M,i} \in X_M$. For building the graph we concatenate the patient metadata $X_M$ and radiomics features $R$ into the numerical features $f_i \in F$ with $f_i = x_{M,i} \concat r_i$ and define the distance between two vertices based on these features as $\omega\left(f_i, f_j\right)$. Each vertex $v_i$ is connected with its $k$ nearest neighbors $v_j \in N_{v_i}$ with the edges $e_{ij} \in E$. $N_{v_i}$ is defined as $N_{v_i} \subseteq V$ s.t. $\left|N_{v_i}\right|=k$ with the non-neighbors $v'$ defined as $\forall{v'} \in V \backslash N_{v_i}$
\begin{equation}
    \operatorname{\omega}\left(v_i, v'\right) \geq \max _{v_j \in N_{v_i}} \operatorname{\omega}\left(v_i, v_j\right).
\end{equation}
For the distance function $\omega$ between two nodes $v_i$ and $v_j$ we choose the weighted Minkowski distance with the order $p$:

\begin{equation}
    \operatorname{\omega}\left(f_i, f_j\right) = \left(\sum_l{\left(\left|w_l \left(f_{i,l} - f_{i,l}\right)\right|^p\right)}\right)^{1/p}.
\end{equation}
 As an alternative to feature selection, we propose to weight each feature based on a statistical analysis of the training data. Statistically important features should therefore have a higher influence on the distance and similarity calculation. Possible weightings include correlation coefficients, e.g. the Pearson correlation for continuous features, or estimated mutual information between the input features and the target labels like $Y_{ICU}$ calculated on the training set. The motivation to use mutual information is to discover non-linear associations between the features and predicted labels, in addition to linear relationships.
 All distances are calculated on the z-scores normalized features.
 In Fig.~\ref{fig:graph_construction} the k-nearest neighbors (KNN) graphs for one training set are visualized with and without weighting of the distance with mutual information.

\subsection{Segmentation and image feature extraction}
\label{sec:segmentation}
The proposed method is built on a joint image feature extraction and segmentation backbone. For this, any encoder-decoder-based architecture with a compressed bottleneck representation and segmentation output can be used. As described with more detail in Sec.~\ref{sec:experiments}, we choose the original 2D U-Net architecture \citep{Ronneberger2015} with small adaptions for our experiments.\\
The $S$ equidistant slices forming an input image $x_{I, i}$ (see Sec.~\ref{sec:equidistant}) are processed as a batch in parallel. Hence, for each slice a 2D segmentation of the healthy lung and pathologies is generated comprising the segmentation output $y_{\text{Seg},i} \in Y_{\text{Seg}}$. The image representation used for the classification task is extracted with a global average pooling of the two-dimensional bottleneck features of each slice, reducing the bottleneck size $c\times d_1 \times d_2$ with the number of channels $c$ and the spatial dimensions $d_1$ and $d_2$ to a vector with the length of $c$ per slice. To transform the resulting $S$ slice-wise image representations to a single patient-wise representation the slice features are then aggregated by taking the element-wise maximum along the stacking dimension resulting in a single vector with size $c$. This vector is then passed through a final fully connected layer followed by a leaky ReLu activation to obtain the latent image representation $z_{I,i} \in Z_{I}$.\\
Based on the improved performance reported by~\cite{goncharov2021ct} using the final feature map of the U-Net instead of the bottleneck, we evaluated this approach, but initial results showed a significant drop in performance which is why we did not investigate this concept any further.

\subsection{Automatic extraction of radiomics}
\label{sec:radiomics_ex}
Inspired by \cite{Burian2020} the patient metadata is complemented with radiomics features $R$ that are automatically extracted from the segmentation output $Y_{\text{Seg}}$. In addition to being more robust to overfitting than extracted image features, this aids the interpretability of the network by providing intermediate results that can easily be verified by visualizing the segmentation output. Both absolute and relative quantification of the COVID-19 burden in the segmented lung are used: The absolute volume of the healthy lung and the two pathology classes (GGO and other pathologies) are calculated by summing up all the predicted voxels per class over each of the slices of the subvolume and multiplying this voxel count with the volume per voxel. Then the total lung volume is calculated by summing up all three classes. The relative pathology burden and ventilation (healthy lung remaining) is calculated by dividing each of the volumes by the total lung volume.

\subsection{Joint feature fusion}
The three input sources provided by the image data $x_{I,i} \in X_I$ and resulting extracted features $z_{I,i}$, extracted radiomics features $r_i \in R$ and metadata $x_{M,i} \in X_M$ constitute three separate modalities used within the graph network to perform the classification task for an individual patient node $v_i$ within the graph. Especially the metadata $X_M$ can provide valuable orthogonal information to the imaging-based other two contributions. To assure that the influence of every modality is equally considered during processing, we are using a linear transformation on every modality to receive a feature representation of equal size. These representations are then processed within an aggregation function $\Psi$ to receive the corresponding combined representation $z_{c,i}$ used within the graph network:
\begin{equation}
    z_{c,i} = \Psi\left(\sigma\left(\Theta_I z_{I,i}\right), \sigma\left(\Theta_R r_i\right), \sigma\left(\Theta_M x_{M,i}\right)\right) ~,
\end{equation}
where $\sigma$ is a non-linear activation function and $\Theta_I  \in \mathbb{R}^{F_I \times F_c}$, $\Theta_R \in \mathbb{R}^{F_R \times F_c}$, $\Theta_M \in \mathbb{R}^{F_M \times F_c}$ are learnable linear transformations, which map the incoming feature dimension onto dimension $F_c$. Possible approaches for $\Psi$ are concatenation, averaging, pooling or attention mechanisms.

\subsection{Classification with GAT}
The graph processing of our proposed method is based on graph attention layers (GAT)~\citep{Velickovic2018}. They combine effective processing of the provided neighborhood with the possibility for direct inference on new unseen data samples while maintaining filter localization and low computational complexity. For more details see Appendix~\ref{app:theory}. Every GAT layer corresponds to the aggregation of a 1-hop neighborhood within graph $G(V,E)$ for every vertex $v_i$. For every vertex representation, not only the representation itself but also neighboring feature vectors are considered for the transformation within a network layer. For GAT, the consideration of every neighbor is conditioned on a learned attention coefficient, which weights every feature vector for its importance in the update step of the representation of interest. The update of the provided combined feature representation $z_{c,i}$ within a GAT layer is based on Eq.~\ref{eq:gat_layer}:
\begin{equation}
    z_{c,i}' = \concat_{p=1}^P \sigma \left( \sum_{j \in N(i)} \alpha_{ij}^p \Theta^p z_{c,j}\right) ~,
\label{eq:gat_layer}
\end{equation}
where $\alpha_{ij}^p$ is the learned attention coefficient for attention head $p$ encoding the importance of $z_{c,j}$ in the 1-hop neighborhood $N(i)$ for the update of $z_{c,i}$ and $\Theta^p$ is a learned linear transformation. Note that $\alpha_{ii}^p$ is the self-attention. A detailed theoretical derivation of the GAT network structure is given in Appendix~\ref{app:theory}.\\
The attention-based graph processing allows us to incorporate the patient metadata $X_M$  effectively into the learning process by basing the graph setup on metadata similarity and creating $N(i)$ for every $z_{c,i}$. Hence, a transformation of representation $z_{c,i}$ does not only rely on the representation itself but receives weighted contributions from all $z_{c,j} \in N(i)$. This process has the potential to stabilize the prediction for patients with an uncharacteristic initial representation of its corresponding class, but which is localized within the correct data cluster.

\begin{table*}[!htp]
    \caption{\label{tab:methods}Backbones and classifiers used for evaluation with the respective features for patients (nodes) and the distance metric (similarity): The proposed U-GAT is the end-to-end trained U-GAT. We compare it to other end-to-end trained methods only using metadata (MLP-Metadata), only using image data (ResNet18) and a GAT with a CNN backbone without an auxiliary segmentation task (ResNet18-GAT). In addition, we compare the performance of different classifiers on the features extracted by a pretrained U-Net, marked with a *, i.e. U-Net*. KNN is a weighted k-nearest neighbors classifier with no weighting of features in the distance metric. wKNN is a KNN classifier using the same weighted distance metric as the proposed graph construction. The MLP classifier is only evaluating the node features without taking the neighbors into account. GraphSAGE is a graph convolutional method without an attention mechanism.}
  \centering
  \vspace{3pt}
  \begin{tabular}{llllllll}
  \toprule
    Architecture & Backbone & Classifier & Node features & Distance features\\
    \midrule
    MLP-Metadata & - & MLP  & Metadata (clinical, no radiomics) & - \\
    ResNet18 & ResNet18 & MLP  & Image repr.  & - \\
    U-Net*+RF & U-Net* & Random Forest  & extr. radiomics + metadata & - \\
    U-Net*+KNN & U-Net* & weighted KNN  & - & extr. radiomics + metadata  \\
    U-Net*+wKNN & U-Net* & weighted KNN  & - & extr. radiomics + metadata  \\
    U-Net*+MLP & U-Net* & MLP  & Image repr. + extr. radiomics + metadata & - \\
    U-Net*+GraphSAGE & U-Net* & GraphSAGE  & Image repr. + extr. radiomics + metadata & extr. radiomics + metadata  \\
    ResNet18-GAT & ResNet18 & GAT   & Image repr. + metadata &  Metadata  \\
    U-GAT* & U-Net* & GAT   & Image repr. + extr. radiomics + metadata & extr. radiomics + metadata  \\
    U-GAT & U-Net & GAT  & Image repr. + extr. radiomics + metadata & extr. radiomics + metadata  \\
    \bottomrule
  \end{tabular}
\end{table*}

\section{Experiments}
\label{sec:experiments}
\subsection{Experimental setup}
We evaluate the proposed method on our in-house dataset using a nested 5-fold cross-validation stratified by the ICU labels. For this, the dataset is split into 5 equally sized folds each containing a similar amount of ICU patients. In nested cross-validation, there are outer and inner evaluation loops for testing and validation. In each of the 5 outer loops, one fold is selected as a test set and the remaining 4 folds are used for training and validation. In the 4 inner loops, 3 folds are selected for training and 1 fold for validation. This is repeated until every combination has been used for testing and validation resulting in a total of 20 repetitions of the experiments. Each repetition is trained with a new seed for the initialization of random states. \\
For the experiments presented here, the lung CT images were used in combination with the following clinical features and blood test results: age, sex, body temperature, percutaneous oxygen saturation, leukocytes, lymphocytes, C-reactive protein (CRP), creatine, D-Dimer, lactate dehydrogenase (LDH), creatine kinase, troponin T, interleukin 6 (IL-6), thrombocytes. The outcomes included: need of mechanical ventilation, admission to ICU and survival of the patient (mortality) as well as the combination of these three outcomes in a multilabel setting.\\
We focus on the evaluation of the main task of ICU prediction, but also extend some experiments on the tasks of ventilation and mortality outcome to explore multitasking and the translation to other tasks.
The experiments were conducted with 10 equidistant samples ($Z=10$) resulting in 9 subvolumes per patient. During training, a random subvolume is chosen for each patient. At validation and test time the whole volume is used since there is only one test patient per batch and the extracted image features and radiomics of the other patients can be loaded.
\subsection{Network parameters and training}
As the segmentation and image feature extraction backbone, we choose the classical 2D U-Net architecture proposed by \cite{Ronneberger2015} with the following modifications in the double convolution blocks: an added batch normalization layer after each activation for faster convergence and a padding of one pixel in each convolution layer to align input and output image size of the network. The final layer consists of a one-dimensional convolution to the number of output classes followed by a softmax layer. The initial filter size of the convolutions is reduced to 32 instead of 64 to accommodate the small input image size of $96\times96$ pixels. We train the model with a batch size of 18 patients each consisting of 10 equidistant slices randomly sampled and the accommodating metadata. A batch normalization and 10\% dropout is applied to the concatenated feature vector $Z$ consisting of image, radiomics and metadata features before being passed to the classification head. Since it drastically decreased the image segmentation performance, we did not backpropagate the classification loss through extracted radiomics $R$ over the U-Net output. We evaluated concatenation, averaging and max pooling for the feature fusion $\Psi$. Following a performance evaluation on the validation set, we received marginally better performance for concatenation without a significant difference compared to the two other approaches. We, therefore, relied on the concatenation approach within our performed experiments.\\
For the graph-based classification head, we choose a GAT with two layers, 5 attention heads and a dropout rate of 10\%. Each node feature vector with an input size of 96 gets refined to a feature size of 64 in the first layer and is reduced to a feature size equal to the number of classification labels in the final node classification layer. All binary classification outputs are finally activated with a sigmoid function. \\
We conducted all experiments in PyTorch 1.7.0 \citep{pytorch} and PyTorch Geometric 1.7.0 \citep{pytorch-geometric} using the Adam optimizer with a base learning rate of $5 \times 10^{-4}$ and a weight decay of $3\times10^{-5}$. All models were trained on an NVIDIA Titan V 12GB GPU using Polyaxon. We define an epoch as 80 patients and train the model for a minimum of 25 epochs in the end-to-end case and for a minimum of 5 epochs when using a pretrained U-Net. Experiments using a pretrained U-Net are indicated with an *, i.e. U-Net*. After the minimum amount of epochs, we stop the training if the validation loss has not improved for 5 epochs. In the end-to-end experiments with joint segmentation and classification, we employ a pretraining schedule since it improved both segmentation and classification results in our preliminary experiments. Here the classification loss was set to zero for the first 20 epochs and the segmentation loss was only trained on the lung masks for the first 10 epochs and then for another 10 epochs on all segmentation labels.
For optimizing the segmentation task we use a Dice loss as introduced by \cite{milletari2016v} and a binary cross-entropy (BCE) loss for all classification tasks.
The total training loss is set to the unweighted sum of the segmentation and classification losses, after both manual and automatic weighting approaches did not yield significant improvements. \\
\textit{The source code to allow easy reproduction of our methodology will be made publicly available upon acceptance of the article.}

\subsection{Data augmentation}
To prevent overfitting on the limited training data, the following augmentation methods were randomly applied to each volume during data loading: intensity scaling with a factor of up to $\pm0.15$, rotation up to $\pm10\degree$ around all axes and an isotropic scaling with a factor between $0.9$ and $1.2$. To always have the same input size, bigger volumes were cropped at a random location and padded with zeros if needed. During validation and testing, the volumes were cropped centrally.

\begin{table*}[!p]
  \caption{\label{tab:mutual_information}Top 10 features sorted by the mutual information for each task and its Pearson correlation. The average is calculated on the training sets of all repetitions. In the multilabel setup, the mutual information with the ordinal regression of outcome severity is estimated for each feature.}
  \centering
  \vspace{3pt}
  \begin{tabular}{lllcc}
  \toprule
    Task & Feature & Category &Mutual information & Pearson correlation\\
    \midrule
    \midrule
    ICU & Healthy lung (\%) & Radiomics & $0.244 \pm 0.052$ & $-0.596 \pm 0.033$\\
    ICU & Ground-glass  opacity (\%) & Radiomics & $0.184 \pm 0.043$ & $+0.577 \pm 0.026$\\
    ICU & Other pathologies (\%) & Radiomics & $0.144 \pm 0.055$ & $+0.471 \pm 0.048$\\
    ICU & C-reactive protein & Clinical & $0.104 \pm 0.038$ & $+0.372 \pm 0.071$\\
    ICU & Interleukin 6 & Clinical & $0.091 \pm 0.023$ & $+0.091 \pm 0.137$\\
    ICU & Age & Clinical & $0.087 \pm 0.031$ & $+0.018 \pm 0.062$\\
    ICU & Lymphocytes & Clinical & $0.047 \pm 0.027$ & $-0.062 \pm 0.112$\\
    ICU & Temperature & Clinical & $0.043 \pm 0.040$ & $-0.016 \pm 0.116$\\
    ICU & Serum creatinine & Clinical & $0.041 \pm 0.045$ & $+0.009 \pm 0.125$\\
    ICU & Thrombocytes & Clinical & $0.039 \pm 0.037$ & $-0.007 \pm 0.060$\\
    ICU & Creatine kinase (total) & Clinical & $0.037 \pm 0.040$ & $+0.113 \pm 0.110$\\
    \midrule
    Ventilation & Healthy lung (\%) & Radiomics & $0.212 \pm 0.033$ & $-0.581 \pm 0.030$\\
    Ventilation & Ground-glass  opacity (\%) & Radiomics & $0.170 \pm 0.022$ & $+0.585 \pm 0.026$\\
    Ventilation & Other pathologies (\%) & Radiomics & $0.159 \pm 0.055$ & $+0.428 \pm 0.051$\\
    Ventilation & Interleukin 6 & Clinical & $0.114 \pm 0.048$ & $+0.109 \pm 0.130$\\
    Ventilation & C-reactive protein & Clinical & $0.082 \pm 0.047$ & $+0.395 \pm 0.070$\\
    Ventilation & Temperature & Clinical & $0.082 \pm 0.044$ & $+0.031 \pm 0.118$\\
    Ventilation & Age & Clinical & $0.059 \pm 0.037$ & $+0.056 \pm 0.053$\\
    Ventilation & Serum creatinine & Clinical & $0.055 \pm 0.034$ & $-0.020 \pm 0.063$\\
    Ventilation & Lactate dehydrogenase  & Clinical & $0.053 \pm 0.028$ & $+0.104 \pm 0.060$\\
    Ventilation & Percutaneous oxygen saturation & Clinical & $0.052 \pm 0.017$ & $-0.285 \pm 0.074$\\
    Ventilation & Creatine kinase (total) & Clinical & $0.045 \pm 0.046$ & $+0.160 \pm 0.106$\\
    \midrule
    Mortality & Healthy lung (\%) & Radiomics & $0.061 \pm 0.040$ & $-0.210 \pm 0.093$\\
    Mortality & C-reactive protein & Clinical & $0.048 \pm 0.034$ & $+0.126 \pm 0.072$\\
    Mortality & Lymphocytes & Clinical & $0.034 \pm 0.040$ & $-0.095 \pm 0.030$\\
    Mortality & Percutaneous oxygen saturation & Clinical & $0.033 \pm 0.038$ & $-0.023 \pm 0.068$\\
    Mortality & Interleukin 6 & Clinical & $0.031 \pm 0.013$ & $+0.068 \pm 0.096$\\
    Mortality & D-dimer & Clinical & $0.030 \pm 0.023$ & $+0.122 \pm 0.117$\\
    Mortality & Temperature & Clinical & $0.022 \pm 0.026$ & $-0.014 \pm 0.068$\\
    Mortality & Lactate dehydrogenase  & Clinical & $0.019 \pm 0.024$ & $+0.246 \pm 0.070$\\
    Mortality & Sex & Clinical & $0.019 \pm 0.010$ & $-0.150 \pm 0.041$\\
    Mortality & Ground-glass  opacity (\%) & Radiomics & $0.018 \pm 0.023$ & $+0.265 \pm 0.079$\\
    Mortality & Other pathologies (\%) & Radiomics & $0.016 \pm 0.024$ & $+0.083 \pm 0.100$\\
    \midrule
    Multilabel & Healthy lung (\%) & Radiomics & $0.274 \pm 0.063$ & $-0.548 \pm 0.051$\\
    Multilabel & Ground-glass  opacity (\%) & Radiomics & $0.190 \pm 0.052$ & $+0.550 \pm 0.042$\\
    Multilabel & Other pathologies (\%) & Radiomics & $0.173 \pm 0.057$ & $+0.407 \pm 0.066$\\
    Multilabel & Interleukin 6 & Clinical & $0.105 \pm 0.040$ & $+0.098 \pm 0.133$\\
    Multilabel & Sex & Clinical & $0.104 \pm 0.110$ & $-0.167 \pm 0.052$\\
    Multilabel & C-reactive protein & Clinical & $0.100 \pm 0.043$ & $+0.352 \pm 0.068$\\
    Multilabel & Lymphocytes & Clinical & $0.062 \pm 0.044$ & $-0.116 \pm 0.046$\\
    Multilabel & Age & Clinical & $0.057 \pm 0.028$ & $+0.057 \pm 0.062$\\
    Multilabel & Percutaneous oxygen saturation & Clinical & $0.047 \pm 0.055$ & $-0.233 \pm 0.069$\\
    Multilabel & Troponin T & Clinical & $0.040 \pm 0.021$ & $+0.068 \pm 0.103$\\
    Multilabel & Temperature & Clinical & $0.036 \pm 0.038$ & $-0.015 \pm 0.108$\\
    \bottomrule
  \end{tabular}
\end{table*}

\subsection{Graph construction}
We selected the KNN graph construction method using a mutual information weighted distance metric for the following experiments by comparing it to other methods on the validation set. For $\omega$ the weighted Euclidean distance (Minkowski distance of second order, $p=2$) is used. The mutual information is estimated using the method proposed by \cite{ross2014mutual} with 3 neighbors averaging the results of 30 repetitions. Using this metric for graph construction was compared to using the Pearson correlation for distance weighting and unweighted KNN graph construction using manually selected features. The evaluated feature sets include all features, clinical metadata, the extracted radiomics and only using age and sex. The number of neighbors $k$ to be considered is set in a hyperparameter search on the validation set. 
Inspired by \cite{parisot2017spectral} the similarity $\operatorname{Sim}(u,v)$ between the two nodes $u, v$ is derived from the distance by applying a radial basis function kernel with the mean distance $\mu$ calculated on the training set:
\begin{equation}
\label{eq:similarity}
    \operatorname{Sim}(u, v) = \exp {\left(-{\frac {\operatorname{\omega}(u,v)} {2\mu ^{2}}}\right)}.
\end{equation}
The scikit-learn library 0.24.1 \citep{scikit-learn} was used for estimating the mutual information and KNN graph construction. SciPy 1.6.2 \citep{scipy} was used for correlation calculation and NumPy 1.18.2 \citep{numpy} for all distance calculations.

\begin{table*}[t!]
  \caption{\label{tab:graph_construction}Evaluation of edge features and weighting used for distance calculation on the validation set. All experiments marked with * have been conducted using a pretrained U-Net. The proposed method of using the estimated mutual information for weighting the features in the distance function is compared to weighting with the Pearson correlation, no weighting and manual feature selection. AP is the average precision and AUC stands for the area under the receiver operating characteristic curve.}
  \centering
  \vspace{3pt}
  \begin{tabular}{lllllll}
  \toprule
    Task & Architecture & Distance features & Distance feature weights & AP & AUC\\
    \midrule \midrule
    ICU & U-GAT*  & age, sex & - &  $0.512 \pm 0.109$ & $0.573 \pm 0.109$ \\
    
    ICU & U-GAT* & clinical& - 
    & $0.671 \pm 0.152$ & $0.720 \pm 0.135$ \\
    ICU & U-GAT* & radiomics & - &  $0.670 \pm 0.145$ & $0.720 \pm 0.116$ \\
    ICU & U-GAT*  & all & - &	$0.704 \pm 0.080$	&	$0.733 \pm 0.073$	\\
    ICU & U-GAT* & all & Pearson correlation &	$0.697 \pm 0.122$	&	$0.751 \pm 0.088$	\\
    ICU & U-GAT* & all & mutual information&	$\textbf{0.722} \pm \textbf{0.096}$	&	$\textbf{0.757} \pm \textbf{0.142}$	\\
    \midrule
    Vent. & U-GAT*& all & - 
    &	$0.640 \pm 0.118$	&	$0.772 \pm 0.072$	\\
    Vent. & U-GAT*& all & mutual information
    &	$\textbf{0.642} \pm \textbf{0.158}$	&	$\textbf{0.793} \pm \textbf{0.114}$	\\
    
    \midrule
    Mort. & U-GAT*& all & - 
    & $\textbf{0.325} \pm \textbf{0.164}$ & $\textbf{0.646} \pm \textbf{0.173}$ \\
    Mort. & U-GAT*& all & mutual information
    &	$0.271 \pm 0.123$	&	$0.588 \pm 0.188$	\\
    \bottomrule

  \end{tabular}
\end{table*}

\subsection{Ablative testing}
In order to understand the effect of the different components of our method, we perform an extensive ablation study on the test set. The main components we evaluate are the image and radiomics feature extraction of the U-Net and the GAT classification. The end-to-end U-Net feature extraction is compared with features from a pretrained U-Net as well as the end-to-end image features from a ResNet18 as proposed by \cite{He2016}. It is important to note that radiomics were not used in the ResNet18-GAT architecture. To evaluate the GAT we replace it with the following classification methods:
\begin{itemize}
    \item Weighted K-nearest neighbors (KNN): 
    The default sklearn weighted k-nearest neighbor classifier using the inverse Euclidean distance of all features as the similarity metric for neighbor selection and for weighting of neighbor labels.
    \item KNN with weighted features (wKNN): The same KNN classifier using the proposed similarity metric (Eq.~\ref{eq:similarity}) with mutual information weighted features. This classifier is using the same graph as the proposed method.
    \item Multilayer Perceptron (MLP): This classifier is a simple neural network with a hidden layer size of 64 followed by a leaky ReLU activation and a 10\% dropout.
    \item GraphSAGE: replacing the GAT operator with GraphSAGE \citep{Hamilton2017}. More details about this method can be found in Appendix~\ref{app:theory}.
\end{itemize}
In addition to ablative testing, the performance of using only clinical metadata or only image features extracted by a ResNet18 is evaluated with an MLP classifier. An overview of the type of data used in each method is given in Table~\ref{tab:methods}.
\subsection{Inner loop ensembles and comparison with Random Forest}
\label{sec:randomforest}
Random Forest is an ensemble method that has been shown to be an effective classifier for small datasets since they do not overfit due to the Law of Large Numbers \citep{Breiman2001} and provide the additional benefit of interpretability. As discussed in Sec.~\ref{sec:intro_feature_fusion} \cite{Burian2020}, \cite{Chao2021} and others have successfully deployed Random Forests to combine extracted image features and metadata for ICU prediction.\\
In this experiment, we focus on the task of ICU prediction and explore if an ensemble of our method can improve its performance and how it compares to the well-established Random Forest classifier. To form an ensemble we average the predicted probabilities of the 4 inner loop models that share the same test set and evaluate the models on the 5 test sets of the outer loop. The Random Forest was implemented using scikit-learn 0.24.1 \citep{scikit-learn} with default parameters.

\subsection{Metrics for segmentation and classification}
As our proposed method follows a multitask approach, the evaluation criteria can be divided into segmentation and classification metrics. \\
For measuring the overlap between segmented regions and the ground truth, we use the Dice score (DS), which is well-established in the medical community. For each class $c$ of background, healthy lung, ground-glass opacity (GGO) and other pathologies, we calculate its Dice score $DS_c$ as
\begin{equation}
    DS_c = \frac{2|P_c \cap G_c|}{|P_c|+|G_c|}
\end{equation}
where $P_c$ is the set of all voxels predicted to be of class $c$, and $G_c$ holds all voxels that are of class $c$ in the segmentation ground truth. If no class is given, the reported DS is the mean of all classes including the background. The reported lung pathology DS is calculated by merging the ground glass opacity and other pathologies into a single label, not by averaging their respective DS.\\
All classification metrics are reported separately for each binary classification task.
The main metrics for evaluating the classification performance are average precision (AP) and the area under the receiver operating characteristic curve (AUC) as they are independent of selected thresholds. 
Given that all tasks have a severe class imbalance, balanced accuracy score (bACC) and F1 score (F1) have been chosen as the main threshold-dependent metrics. In the ensemble experiments, the sensitivity and specificity are additionally reported. For all threshold-dependent metrics, the optimal threshold is set using the validation results and maximizing the Youden's J statistic \citep{youden1950index}:

\begin{equation}
    {J={\text{sensitivity}}+{\text{specificity}}-1}.
\end{equation}
All metrics based on optimized thresholds are indicated with **, e.g. F1**. The classification metrics were calculated using scikit learn 0.24.1 \citep{scikit-learn}.

\begin{table*}[!htp]
  \caption{\label{tab:ablative_testing}Ablative testing and comparison with an MLP only using clinical metadata and a ResNet18 only using image data as input on all tasks. We report the average precision (AP), the area under the receiver operating characteristic curve (AUC) as well as the balanced accuracy (bACC) and F1 score. Metrics calculated on Youdan's J optimized thresholds are marked with **. The variation U-GAT* refers to the proposed method using image and radiomic features extracted from a pretrained U-Net.}
  \centering
  \vspace{3pt}
  \begin{tabular}{lllllll}
  \toprule
    Task & Architecture & AP & AUC & bACC** & F1** \\
    \midrule \midrule
    ICU & MLP-Metadata 	&	$0.577 \pm 0.109$	&	$0.654 \pm 0.104$	&	$0.577 \pm 0.087$	&	$0.560 \pm 0.107$\\
    ICU & ResNet18 	&	$0.670 \pm 0.097$	&	$0.716 \pm 0.077$	&	$0.617 \pm 0.063$	&	$0.560 \pm 0.084$\\
    ICU & U-Net*+KNN	
    &	$0.632 \pm 0.113$	&	$0.677 \pm 0.112$	&	$0.589 \pm 0.087$	&	$0.519 \pm 0.131$\\
    ICU & U-Net*+wKNN	&	$0.659 \pm 0.107$	&	$0.735 \pm 0.087$	&	$0.630 \pm 0.098$	&	$0.525 \pm 0.194$	\\
    ICU & U-Net*+MLP 	&	$0.615 \pm 0.127$	&	$0.687 \pm 0.128$	&	$0.630 \pm 0.098$	&	$0.612 \pm 0.085$	\\
    ICU & U-Net*+GraphSAGE
    &	$0.628 \pm 0.114$	&	$0.690 \pm 0.107$	&	$0.623 \pm 0.079$	&	$0.574 \pm 0.085$
    \\
    ICU & ResNet18-GAT 	&	$0.637 \pm 0.165$	&	$0.678 \pm 0.160$	&	$0.619 \pm 0.097$	&	$0.595 \pm 0.084$\\
    ICU & U-GAT* 	&	
    $0.672 \pm 0.129$	&	$0.725 \pm 0.107$	&	$\textbf{0.690} \pm \textbf{0.094}$	&	$0.651 \pm 0.104$ \\
    ICU + Seg. & U-GAT 	&	$\textbf{0.699} \pm \textbf{0.149}$	&	$\textbf{0.743} \pm \textbf{0.103}$	&	$0.687 \pm 0.095$	&	$\textbf{0.661} \pm \textbf{0.084}$\\
    \midrule
    Ventilation & MLP-Metadata 	&	$0.527 \pm 0.167$	&	$0.692 \pm 0.109$	&	$0.604 \pm 0.124$	&	$0.475 \pm 0.188$\\
    Ventilation & ResNet18 	&	$0.573 \pm 0.127$	&	$0.715 \pm 0.086$	&	$0.601 \pm 0.066$	&	$0.390 \pm 0.160$\\
    Ventilation & U-Net*+KNN 	
   &	$0.527 \pm 0.180$	&	$0.674 \pm 0.112$	&	$0.583 \pm 0.101$	&	$0.368 \pm 0.192$\\
    Ventilation & U-Net*+wKNN 	
    &	$0.591 \pm 0.138$	&	$0.765 \pm 0.071$	&	$0.655 \pm 0.102$	&	$0.453 \pm 0.206$\\
    Ventilation & U-Net*+MLP 	&	$0.587 \pm 0.183$	&	$0.741 \pm 0.119$	&	$0.651 \pm 0.073$	&	$0.488 \pm 0.134$	\\
    Ventilation & U-Net*+GraphSAGE
    &	$0.603 \pm 0.151$	&	$0.758 \pm 0.109$	&	$0.666 \pm 0.115$	&	$0.481 \pm 0.205$
    \\
    Ventilation & ResNet18-GAT 	&	$0.570 \pm 0.152$	&	$0.689 \pm 0.152$	&	$0.598 \pm 0.107$	&	$0.423 \pm 0.178$\\
    Ventilation & U-GAT* 	&	$0.618 \pm 0.137$	&	$\textbf{0.788} \pm \textbf{0.106}$	&	$\textbf{0.720} \pm \textbf{0.120}$	&	$\textbf{0.592} \pm \textbf{0.130}$	\\
    Vent. + Seg. & U-GAT 	&	$\textbf{0.644} \pm \textbf{0.142}$	&	$\textbf{0.788} \pm \textbf{0.112}$	&	$0.699 \pm 0.104$	&	$0.539 \pm 0.179$\\
    \midrule
    Mortality & MLP-Metadata  	&	$0.261 \pm 0.135$	&	$0.544 \pm 0.134$	&	$\textbf{0.551} \pm \textbf{0.080}$	&	$0.224 \pm 0.152$\\
    Mortality & ResNet18  	&	$0.210 \pm 0.116$	&	$0.461 \pm 0.155$	&	$0.493 \pm 0.072$	&	$0.155 \pm 0.138$\\
    Mortality & U-Net*+KNN 
    &	$0.257 \pm 0.137$	&	$0.512 \pm 0.166$	&	$0.513 \pm 0.085$	&	$0.184 \pm 0.147$
    \\
    Mortality & U-Net*+wKNN 	&	$0.214 \pm 0.146$	&	$0.437 \pm 0.164$	&	$0.496 \pm 0.107$	&	$0.171 \pm 0.143$	\\
    Mortality & U-Net*+MLP 	&	$0.252 \pm 0.157$	&	$0.502 \pm 0.191$	&	$0.504 \pm 0.122$	&	$0.190 \pm 0.157$	\\
    Mortality &U-Net*+GraphSAGE
    &	$0.270 \pm 0.143$	&	$0.568 \pm 0.180$	&	$0.545 \pm 0.139$	&	$0.236 \pm 0.163$		
    \\
    Mortality &ResNet18-GAT 	&	$0.247 \pm 0.151$	&	$0.520 \pm 0.156$	&	$0.490 \pm 0.154$	&	$0.184 \pm 0.157$\\
    Mortality &U-GAT* 	&	$0.271 \pm 0.137$	&	$0.549 \pm 0.188$	&	$0.529 \pm 0.133$	&	$\textbf{0.230} \pm \textbf{0.172}$	\\
    Mort. + Seg. & U-GAT 	&	$\textbf{0.287} \pm \textbf{0.186}$	&	$\textbf{0.586} \pm \textbf{0.187}$	&	$0.532 \pm 0.157$	&	$0.199 \pm 0.173$\\
    \bottomrule
  \end{tabular}
\end{table*}

\begin{figure*}[!htp]
    \centering
    \includegraphics[width=0.92\textwidth]{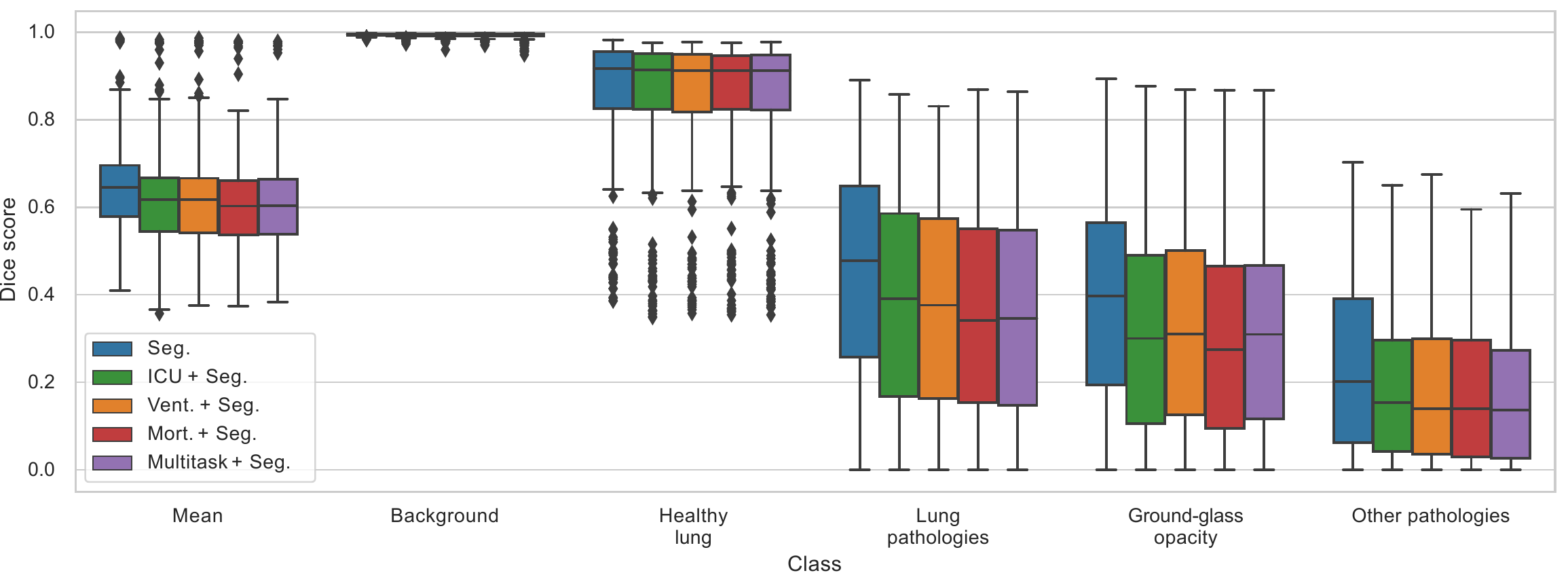}
    \caption{\label{fig:dice}Dice scores of the different multitasking approaches compared with the single task segmentation. While the auxiliary segmentation task improves classification results, the segmentation performance is lower in all multitask settings in comparison to the U-Net only optimized for segmentation.}
\end{figure*}

\begin{table*}[]

    \caption{Results for the multitasking of the segmentation of pathologies and the prediction of all three patient outcomes ICU admission, need for ventilation and mortality. The graph construction is based on the ordinal regression of outcome severity. Each outcome prediction is modelled as a non-exclusive binary classification i.e. a multilabel problem. The mortality task is the only task benefiting from this multitask setup.}
    \label{tab:multilabel}
    \centering
    \begin{tabular}{lllllll}
    \toprule
    Task & Architecture & AP & AUC & bACC** & F1** \\
        \midrule
        ICU & U-GAT 	&	$0.649 \pm 0.128$	&	$0.697 \pm 0.116$	&	$0.642 \pm 0.097$	&	$0.569 \pm 0.163$\\
        Ventilation & U-GAT 	&	$0.622 \pm 0.127$	&	$0.774 \pm 0.094$	&	$0.681 \pm 0.102$	&	$0.503 \pm 0.188$\\
        Mortality & U-GAT 	&	$0.289 \pm 0.138$	&	$0.620 \pm 0.175$	&	$0.536 \pm 0.133$	&	$0.216 \pm 0.174$\\
    \bottomrule
    \end{tabular}
\end{table*}

\section{Results}
\label{sec:results}
\subsection{Graph construction method}
In the first part of the experiments, we try to find the optimal graph construction method by evaluating different features and distance weights used in the KNN graph construction on the validation set. Using a simple, unweighted KNN classifier, the impact of the number of neighbors was evaluated on the ICU task using all features. A hyperparameter search using $k \in \{3,5,7,9, 11\}$ concluded that connecting each node with its seven nearest neighbors yields the best results and $k=7$ was therefore used for all graph construction experiments.
Both mutual information and the Pearson correlation were evaluated for weighting features in the distance calculation. Table~\ref{tab:mutual_information} shows the top 10 of the average of both measures throughout all folds, sorted by mutual information. While a Pearson correlation $>0.3$ and mutual information $>0.1$ can be observed in the ICU, ventilation and multilabel tasks for some features, the mortality label shows lower mutual information and correlation on all features. The percentage of the healthy lung has the highest mutual information for all tasks. To avoid test set leakage, the mutual information was calculated at the beginning of each run on the respective training set only.\\
The results for the different graph construction methods are shown in Table~\ref{tab:graph_construction}. Only using age and sex as features for graph construction performed the worst. While considering only clinical metadata or only radiomics showed comparable results, using all available features outperformed both. Using the Pearson correlation to weight features increased the AUC, but lowered the AP. The proposed weighting with mutual information performed best for the ICU task with an AP of $0.722\pm0.096$ and an AUC of $0.757\pm0.142$. The mutual information weighted KNN graph construction method was also compared to the baseline using all features without weighting in the tasks of ventilation and mortality prediction. While performing better for the ventilation task, it failed to outperform the baseline for the mortality~task~(Table~\ref{tab:graph_construction}). \\
A hyperparameter search for $k$ using the proposed method on the ICU task confirmed the choice of $k=7$ and was thus used for all subsequent experiments.
To mitigate the heavy class imbalance of the segmentation task we use the ground truth values of all radiomics for the graph construction of all training samples and only use the predicted radiomics on the test and validation patients to connect them to the graph.

\subsection{Ablative testing}

In the next set of experiments, we evaluate the different components of the proposed method and compare the results to baseline methods.
Table~\ref{tab:ablative_testing} shows the results of ablative testing and comparison with an MLP only using metadata and a ResNet18 only using image data. For both, the ICU task and the ventilation task, the MLP model has the lowest AP. In contrast, in the mortality task, only the graph-based U-Net architectures have a higher AP. It can also be observed, that the image-based ResNet18 has the lowest AP for mortality prediction. For both the ICU and ventilation tasks, the ResNet18 has a lower AP than the proposed method. All ablations of replacing the U-Net with a ResNet18 and replacing the GAT with an MLP or a GraphSAGE result in a lower AP in all tasks. \\
The results of joint end-to-end training of the segmentation and classification task seem to improve the AP for all tasks slightly. However, the average Dice score is lower in all multitask setups than in the segmentation single task setup (see Fig.~\ref{fig:dice}). In this setup, both the ICU prediction and the ventilation prediction reached the highest AP of $0.699\pm0.149$ and $0.644\pm0.142$, respectively. The end-to-end multitasking of the classification of all labels and segmentation only benefited the mortality task with the highest AP of $0.289\pm0.138$ and AUC of $0.620\pm0.175$ in this setup (see Table \ref{tab:multilabel}).

\subsection{Inner loop ensembles and comparison with Random Forest}
\begin{figure*}[!t]
    \centering
    \includegraphics[width=\textwidth]{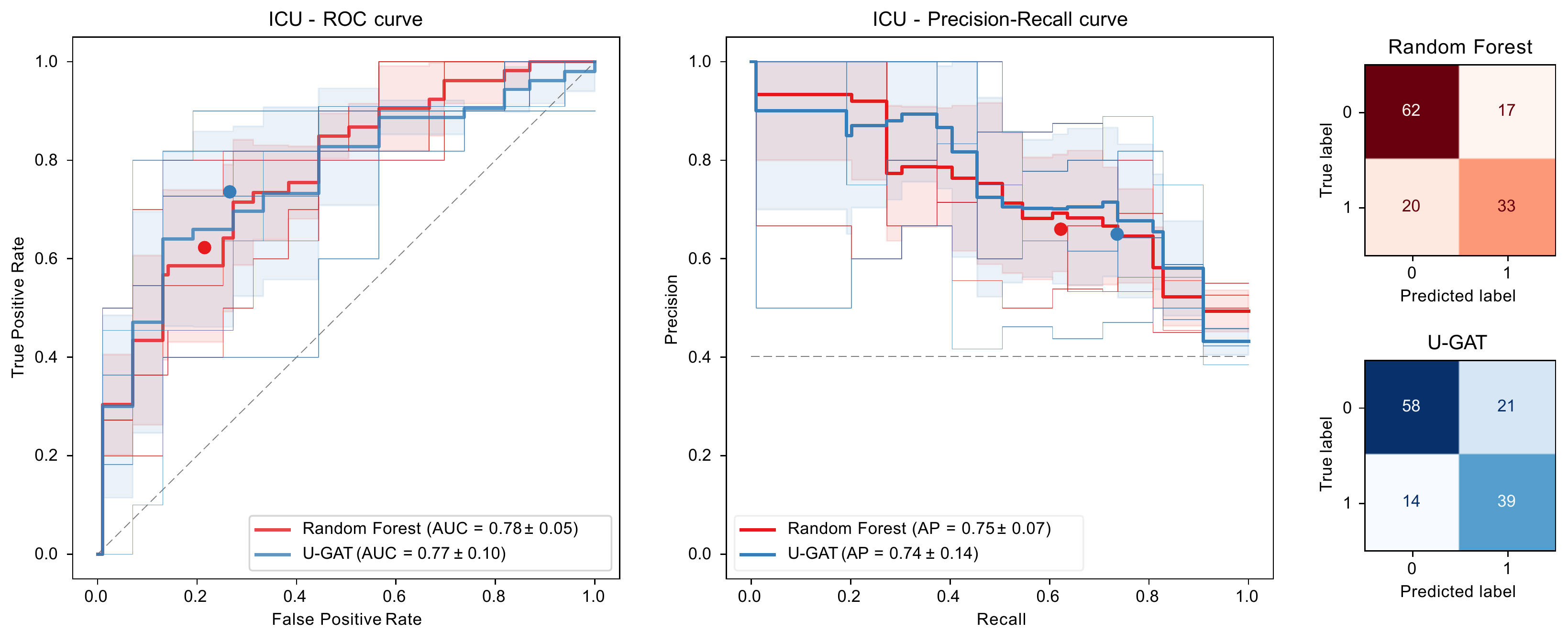}
    \caption{Ensemble results for the ICU task: receiver operating characteristic (ROC) and precision-recall curve comparing inner cross-validation loop ensembles of Random Forest and the proposed U-GAT method. Each narrow curve represents the results of one of the 5 test sets. The average of these is estimated with the bold curve and its standard deviation is depicted by the shaded area. The confusion matrices on the right include the predictions for all patients comprising all 5 test folds with optimized thresholds. The corresponding metrics of these confusion matrices are visualized with filled circles in the curve diagrams.}
    \label{fig:roc}
\end{figure*}

As discussed in Sec.~\ref{sec:randomforest}, we also perform a comparison of our method against Random Forests which have been used in previous works. The ICU task results with and without ensembles and comparison with Random Forests are shown in Table~\ref{tab:results_ensemble}. Ensembling increases the AP from $0.729\pm0.089$ to $0.751\pm0.067$ for the Random Forest and from $0.699\pm0.149$ to $0.745\pm0.137$ for the U-GAT, resulting in a $0.024$ higher increase in AP by ensembling for the U-GAT. Both AP and AUC are higher in the respective RF models. However, the balanced accuracy, F1 score and sensitivity are highest for the U-GAT ensemble, while specificity is highest for the Random Forest ensemble. The standard deviations and the receiver operating characteristic (ROC) and precision-recall curves in Fig.~\ref{fig:roc} show that while the averages of both ensembles are close to each other, the spread of the U-GAT is higher with outliers in both directions. Using a Wilcoxon signed-rank test we could not show a significant difference in AP or AUC between the two ensembles.

\begin{table*}[!t]
  \caption{\label{tab:results_ensemble}ICU task results of a Random Forest (RF) and the proposed U-GAT with and without inner cross-validation loop ensembles. In addition to the metrics used in the ablation study, we report the sensitivity and specificity. Metrics calculated on Youdan's J optimized thresholds are marked with **.}
  \centering
  \vspace{3pt}
    \begin{tabular}{lllllll}
    \toprule
        Architecture	&	 AP	&	 AUC	&	 bACC**	&	 F1**	&	 Sens.**	&	 Spec.**	\\
        \midrule
        U-Net*+RF 	&	$0.729 \pm 0.089$	&	$0.774 \pm 0.057$	&	$0.716 \pm 0.075$	&	$0.649 \pm 0.011$	&	$0.651 \pm 0.177$	&	$0.781 \pm 0.166$	 \\
        U-Net*+RF ensemble 	&	$\textbf{0.751} \pm \textbf{0.067}$	&	$\textbf{0.781} \pm \textbf{0.046}$	&	$0.708 \pm 0.083$	&	$0.628 \pm 0.135$	&	$0.631 \pm 0.218$	&	$\textbf{0.784} \pm \textbf{0.157}$	 \\
        U-GAT 	&	$0.699 \pm 0.149$	&	$0.743 \pm 0.103$	&	$0.699 \pm 0.852$	&	$0.647 \pm 0.097$	&	$0.670 \pm 0.115$	&	$0.728 \pm 0.141$	 \\
        U-GAT ensemble 	&	$0.745 \pm 0.137$	&	$0.770 \pm 0.098$	&	$\textbf{0.735} \pm \textbf{0.111}$	&	$\textbf{0.700} \pm \textbf{0.114}$	&	$\textbf{0.736} \pm \textbf{0.067}$	&	$0.734 \pm 0.174$	 \\
        \bottomrule
    \end{tabular}
\end{table*}

\section{Discussion}
\label{sec:discussion}
\begin{figure*}[ht!]
    \centering
    \includegraphics[trim=90 360 80 70,clip,width=\textwidth]{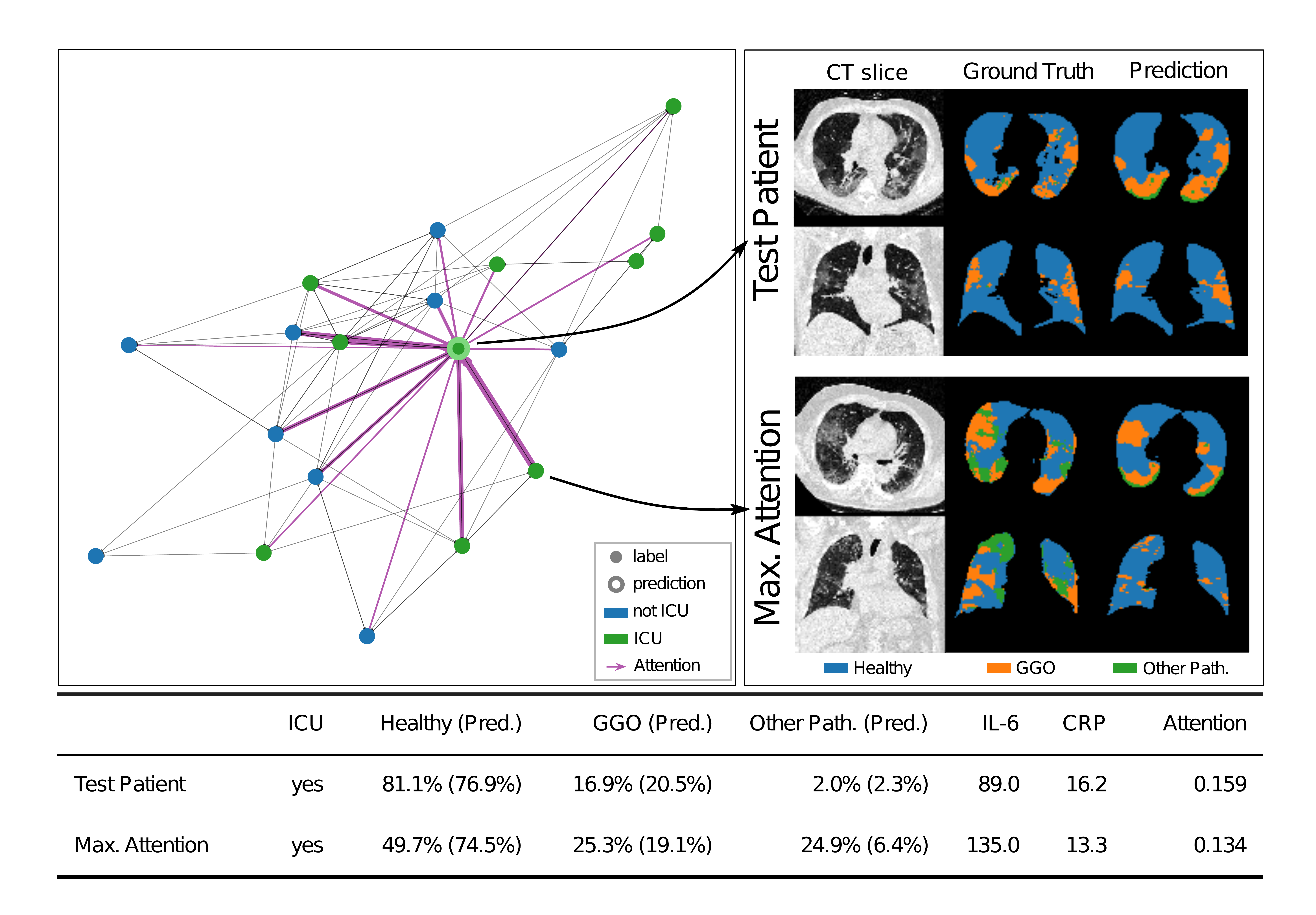}
    \begin{tabularx}{\textwidth}{llllllll}
    \toprule
                     & ICU & Healthy (Prediction) & GGO (Prediction) & Other Path. (Prediction) & IL-6 & CRP & Attention \\
    \midrule
        Test Patient & yes & 81.1\% (76.9\%) & 16.9\% (20.5\%) & 2.0\% (2.3\%) & 89.0 & 16.2 & 0.159 \\
        Max. Attention& yes & 49.7\% (74.5\%) & 25.3\% (19.1\%) & 24.9\% (6.4\%) & 135.0 & 13.3 & 0.134 \\
        \bottomrule
    \end{tabularx}
    \caption{Left: Batch graph showing the attention scores of a single test patient. The thickness of the line corresponds to the attention score of the respective neighbors after two hops. Right: CT images, segmentation ground truth and predicted segmentation of a single axial and coronal slice from the test patient and the neighbor with maximum attention. Bottom: Most important features for the test patient and the neighbor with maximum attention. In brackets, the  radiomics predicted by the pretrained U-Net are shown. }
    \label{fig:attention}
\end{figure*}

\subsection{The synergistic effect of multitasking}

The different multitasking setups of the proposed method were evaluated to investigate if any synergistic effects can improve the generalization of the models. In comparison to a U-Net only trained for segmentation, any addition of classification tasks decreased the performance for the segmentation task. Also, the hypothesis that the segmentation of underrepresented classes might be improved by multitasking could not be confirmed. In contrast to this, a joint segmentation seems to improve the classification performance of all classification tasks, which makes the segmentation task a suitable auxiliary task to improve classification results. This could clearly be observed when comparing the GAT classification of image features extracted by a ResNet18 with the bottleneck features extracted from the U-Net. However, this difference might also be attributed to the U-Net providing radiomics for the graph construction and as patient features. The synergistic effect of multitasking and end-to-end image extraction could be shown for all tasks with the further improved results of U-GAT over the sequentially trained U-GAT* with fixed bottleneck features.\\
Predicting the labels of all classification tasks at the same time showed slightly improved results only for the mortality prediction. This task has the highest class imbalance and also the lowest mutual information with the clinical metadata and radiomics. This indicates that the joint prediction of less severe outcomes as an auxiliary task might improve the challenging prediction of mortality. 

\subsection{Patient similarity and graph construction}
The increase in performance of all graph-based methods over the MLP baseline shows that leveraging similar patients from the training set is useful for refining the features of test patients. However, the results also show that graph construction is particularly critical for the effect of feature refinement. When using just age and sex, the AP dropped by almost $0.2$ compared to simply using all features when predicting the ICU label. In contrast, the clinically motivated combination of extracted radiomics with metadata for graph construction was reaffirmed by an increase in performance, both over using only metadata and only radiomics.\\
In all tasks the radiomic feature \textit{percentage of healthy lung remaining} had the highest mutual information, giving it the highest weight for graph construction. Both ICU and ventilation labels had the highest mutual information with the three radiomics. Contrary to this, the mutual information and correlation of all features with the mortality outcome was very low and the graph construction method based on this did not show to be effective. However, the weighted graph based on the ordinal multilabel yielded the highest performance for the mortality task indicating that the ICU and ventilation labels might be useful auxiliaries for constructing a graph for the mortality prediction.\\
A key benefit of using a weighted distance for KNN graph construction is that the graph can adapt to each task without prior feature selection. Fig.~\ref{fig:graph_construction} shows the graph for each task with and without weighting the distance measure with mutual information.
Besides improving classification, an effective similarity measure can be used to identify relevant patients that have been treated in the past and support the decision-making process of physicians by enabling them to analyse the disease progression of similar patients.

\subsection{Interpretability and graph-attention}
In addition to its performance boost over comparable GNN methods, using GAT offers another important advantage. The attention mechanism that learns to identify neighbors in the graph that are most relevant for the prediction task of an individual node provides insight into the decision process of the model. The analysis of attention scores could suggest patients that the model deems relevant for the outcome prediction. These connections within the patient population can uncover new information regarding a novel pandemic and provide valuable information to clinicians.\\
During the testing phase, a batch graph consists of one test node and 18 neighboring nodes from the training set that serve as a context for this new patient. For each of the two GAT layers, the model assigns attention scores to the neighbors of each node in the graph, deciding how much the representation of a node after the layer will be based on the representation of its specific one-hop neighbors. These attention scores can be thought of as a weighted directed adjacency matrix $A\in[0,1]^{N\times N}$, where $N$ is the number of nodes in the batch and all rows in $A$ add up to $1$. We can multiply the attention matrices of both layers to receive a matrix that shows how the representation of a node is based on its two-hop neighborhood. These attention scores are visualized in Fig.~\ref{fig:attention}. \\
It highlights that the attention mechanism succeeds in assigning high importance to neighbors of the same class and lower importance to those of the opposite class, thus implicitly refining the neighborhood constructed by the KNN algorithm. Furthermore, we can see that the attention mechanism does not necessarily assign high attention to neighbors that are particularly similar in their features. In contrast to a simple KNN classifier, that can only base its prediction on feature similarity, our method evidently learns complex functions to identify the most relevant neighbors and thus introduces orthogonal information to that embedded in the KNN graph.

\subsection{Challenges and future outlook}
Our in-house dataset offers a unique combination of extensive metadata, various patient outcomes and fine-grained annotated CT images. Acquiring and curating such a dataset requires a large amount of time and resources, thus it was crucial to propose an approach that would not be severely challenged by our limited amount of patients. However, to enhance the results further and ensure generalizability to multiple sites and waves of COVID-19 patients it would be crucial to acquire a larger dataset.
Specifically, a current limitation of our model is the segmentation of underrepresented lung pathologies, such as pleural effusion, as well as the prediction of imbalanced outcomes like mortality. Both of those shortcomings could be overcome by utilizing more annotated data in the future and a wider patient group.\\
We also observed that weighting the distance function with the mutual information of each feature improves the graph structure and the features elevated by this are also in alignment with recent radiological findings~\citep{Burian2020, Chao2021, gong2021multi}. However, the mutual information showed a large standard deviation and was particularly low for the mortality task, highlighting the difficulty of this task and the limitation of the graph construction method for tasks with low mutual information. This again emphasizes the need for a more balanced, bigger dataset.

\section{Conclusions}
\label{sec:conclusion}
In this work, we developed and evaluated a method to effectively leverage multimodal information for the outcome prediction of COVID-19 patients. Here, the said information in form of CT lung scans, patient medical and meta information was incorporated into a graph structure and processed within a GCN to stabilize and support the prediction based on data similarity. With the GCN, we propose an end-to-end methodology that segments patient COVID-19 pathologies in the lung CT images and uses the combination of imaging and non-imaging data to predict various outcomes. We explicitly incorporate automatically extracted lung radiomics in our architecture and demonstrate an increase in performance. This procedure provides a substantial advantage in fast decision-making since our methodology does not require previous processing of the CT scans by a medical expert to provide these radiomics. Due to the end-to-end approach, here the segmentation of the lung CT scans is used as an auxiliary task oriented towards the final patient outcome classification instead of a stand-alone procedure. We show that the segmentation of pathological changes in the lung due to COVID-19 indeed improves outcome prediction, although the segmentation itself is not improved by combining the tasks within the training process. Further, the end-to-end processing of the data assures that the learned feature representations are meaningful with respect to the graph processing in the GCN.\\
To create the patient graph structure, we propose a weighted graph construction based on mutual information combined with the decisive power of the attention approach of GAT. We were able to show that the performed weighting process resulted in more meaningful graphs which also benefited most of the corresponding prediction tasks. Considering the limited amount of patients available, this stabilizing factor is especially crucial. Further, the added transparency that our attention analysis provides could enable clinicians to gain more trust in the predictions of our approach and help them identify similar and relevant previous patients and thus consider their treatment courses in the assessment and therapy planning for the patient in question.\\
Overall, we were able to show that our developed methodology effectively leverages the information provided by the multimodal data of the examined COVID-19 patients and can support the treatment of patients within the uncertain and difficult situation of a pandemic resulting from a previously unknown disease. In a broader picture, our developed methodology therefore can also provide guidance on how the challenging task of multimodal data processing can be optimized and become more interpretable in these situations.\\
For future work, an evaluation on more annotated patients would allow a better analysis of the generalizability of our methodology. Also, access to datasets for other medical conditions which provide a similar structure of multimodal data could help in the ongoing analysis of our methodology. Here, the graph construction procedure and its impact could be further evaluated. At the same time, a larger dataset would benefit the segmentation of underrepresented pathologies, potentially increasing the information density extractable from the radiomics, further advancing our methodology and possibly having a clinical impact.
\section*{Acknowledgments}
The authors will like to thank the team at the Radiology Department at Klinikum rechts der Isar, in particular Matthias Zierhut for his help annotating CTs and Friedericke Jungmann for her help collecting patient metadata. Further, the authors acknowledge the critical views and feedback provided by Anees Kazi, Roger Soberanis-Mukul, Gerome Vivar and Ahmad Ahmadi, the first two from the Chair for Computer Aided Medical Procedures and Augmented Reality at Technische Universität M\"unchen and the latter two from the German Center for Vertigo and Balance Disorders at Ludwig-Maximilians Universität M\"unchen.\\
The CT annotation and segmentation work was partially funded by EIT Health's rapid response project 20882 "FastRAi". The developments on Graph Convolutional Networks were covered by the Bavarian Research Foundation (BFS) grant AZ-1429-20C.

\appendix
\section*{Appendix}
\section{Theoretical background on graph neural networks}
\label{app:theory}
The following section provides a detailed theoretical background on the underlying concepts of graph neural networks and motivates our design choices described within Sec.~\ref{sec:methodology}. 

\subsection*{Graph Convolutional Networks}
We are assuming a number $N$ of vertices $V$ on a graph with feature representations $X \in \mathbb{R}^{N \times M}$, where M is the dimension of the feature representation on every vertex, and corresponding edges $E$ with weights $W$. Every vertex $v_i \in V$ holds its corresponding feature representation $x_i \in X$. The vertices are connected by a graph $G(V, E, W)$, where an edge $e_{ij} \in E$ with weight $w_{ij} \in W$ corresponds to a connection between vertices $v_i$ and $v_j$. The existing edges $E$ are therefore represented in the weighted adjacency matrix $W \in \mathbb{R}^{N \times N}$.\\
We want to incorporate the graph structure for the processing of signal $X$. To aggregate representations from different vertices, a convolution is a suitable approach. However, a convolution operation on a graph is more complex than a standard convolution on e.g. an image due to its non-Euclidian structure. There has been a body of work in recent years to address this problem in the frame of Graph Convolutional Networks. We will first address the initial spectral approach operating in the Fourier domain to provide a detailed context of the methodology since also some benchmark methods we compare against are based on this concept. After discussing some of the drawbacks of this approach, we will go into detail on the spatial methods on which also our method is based.

\subsubsection*{Spectral Convolution on Graphs}
By theory, the convolution operation can be described as a linear operator in the Fourier domain, that diagonalizes in the eigenbasis of the Fourier space. By transferring the graph signal $X$ into the spectral domain, we can therefore simplify the convolution operation~\citep{Hammond2011}. \\
For a graph structure, the eigenbasis is represented by the graph Laplacian $L$. It can be calculated from the weighted adjacency matrix $W$ as $L=D-W \in \mathbb{R}^{N \times N}$, where $D \in \mathbb{R}^{N \times N}$ is the degree matrix defined by $D_{ii} = \sum_j W_{ij}$ and $D_{ij} = 0,~ \forall~ i \neq j$. We can normalize $L$ by using $\hat{L} = I_n - D^{-1/2} W D^{-1/2}$ with $I_n$ as the identity matrix. The eigenbasis of the graph is represented by the eigenvectors of the normalized graph Laplacian $\hat{L}$. $\hat{L}$ is a real symmetric positive semidefinite matrix, it therefore provides a complete set of eigenvectors $\{u_l\}_{l=0}^{N-1}$ and corresponding eigenvalues $\{\lambda_l\}_{l=0}^{N-1}$. We can define $\hat{L} = U \Lambda U^T$ with $U = [u_0,...,u_{N-1}] \in \mathbb{R}^{N \times N}$ and $U = \text{diag}([\lambda_0,...,\lambda_{N-1}]) \in \mathbb{R}^{N \times N}$. This definition results in the Fourier transform of graph signal $X$ to be $\hat{X} = U^T X \in \mathbb{R}^{N \times M}$ as well as the reverse transform $X = U \hat{X}$~\citep{Defferrard2016}.\\
We now can define a learnable linear translation operator and apply it to the graph signal in the Fourier domain. We define this translation operator as $g_\theta$. The convolution operation in Fourier space can therefore be described as follows:
\begin{equation}
    g_\theta \ast X = U g_\theta U^T X
\label{eq:graph_conv}
\end{equation}
The translation operator $g_\theta$ can be learned by applying it to the signal X in Fourier space and transferring the resulting representation into the spatial domain. Since $\hat{L} = U \Lambda U^T$, it further becomes clear that $g_\theta$ can be seen as a function of the Laplacian eigenvalues lambda $g_\theta(\Lambda) = \text{diag}(\theta)$. This corresponds to a non-parametric filter, where $\theta \in \mathbb{R}^{N}$ represents the Fourier coefficients. However, a non-parametric filter $g_\theta(\Lambda)$ has a learning complexity of $\mathcal{O}(N)$. Additionally, a transformation of the signal into Fourier domain and back follows a complexity of $\mathcal{O}(N^2)$. Since the spectral signal in Fourier domain is defined by the whole graph signal $X$, the approach would further normally not result in locality of the learned filters. By using an effective parameterization, \cite{Defferrard2016} however, overcome these problems.\\
From $\hat{L}^k = (U \Lambda U^T)^k = U \Lambda^k U^T$, it follows that $g_\theta(\Lambda)$ can be written as a function of $g_\theta(\hat{L})$. Following~\cite{Hammond2011}, we describe $g_\theta(\hat{L})$ as a polynomial computable from $\hat{L}$ in a recursion in terms of Chebyshev polynomials. We can recursively define them as follows:
\begin{equation}
    T_k(x) =
\begin{cases}
    1,& \text{if } k = 0 \\
    x,& \text{if } k = 1 \\
    2 x T_{k-1}(x) - T_{k-2}(x),& \text{if } k > 1
\end{cases}
\label{eq:cheb_polynom}
\end{equation}
Our filter of localization $K$ can be parameterized by the following Chebyshev polynomial up to the $K^\text{th}$ order:
\begin{equation}
    g_{\theta'}(\Lambda) \approx \sum_{k=0}^K \theta'_k T_k(\Bar{\Lambda})
\label{eq:cheb_approx}
\end{equation}
where $\Bar{\Lambda} = \frac{2}{\lambda_{max}} \Lambda - I_N$ and $\lambda_{max}$ corresponds to the largest eigenvalue of $\hat{L}$. $\theta' \in \mathbb{R}^K$ refers to a vector of Chebyshev coefficients. $\Lambda$ can be substituted by $\hat{L}$ within Eq. \ref{eq:cheb_approx} due to the above mentioned relation $\hat{L}^k = U \Lambda^k U^T$.\\
\cite{Hammond2011} proved that $(\hat{L}^k)_{ij} = 0$ when the path length $d_G$ within the graph is longer than $k$. Here, the path length refers to the minimum number of edges that need to be passed to travel from vertex $v_i$ to $v_j$. Hence, Eq. \ref{eq:cheb_approx} automatically results in a K-localization for a filter updating vertex $v_i$, which can be expressed in:
\begin{equation}
    (\hat{L}^k)_{ij} = 
\begin{cases}
    B,& \text{if } d_G(i,j) \leq k\\
    0,              & \text{otherwise}
\end{cases}
\label{eq:locality}
\end{equation}
where $B$ is a non-zero entry of the $k^\text{th}$ polynomial component. A polynomial of degree $K-1$ containing $(\hat{L}^{K-1})_{ij}$ updates vertex $v_i$ with representation $x_i$ based on the $(K-1)$-hop neighborhood, in which it is localized.
We can therefore define the convolution of the graph signal $X$ with filter $g_\theta'$:
\begin{equation}
    g_{\theta'} \ast X \approx \sum_{k=0}^K \theta'_k T_k(\Bar{L}) X ~,
\label{eq:cheb_conv}
\end{equation}
where $\Bar{L} = \frac{2}{\lambda_{max}} \hat{L} - I_N$. For a sparse Laplacian this operation follows a complexity of $\mathcal{O}(K|E|) \ll \mathcal{O}(N^2)$. A graph convolutional layer using Eq. \ref{eq:cheb_conv} can be defined based on this finding as shown in Eq. \ref{eq:cheb_layer}:
\begin{equation}
    X' = \sigma \left( \sum_{k=0}^K \theta'_k T_k(\Bar{L}) X \right) ~,
\label{eq:cheb_layer}
\end{equation}
where $\sigma$ is a non-linear activation function and $X'$ corresponds to the new feature representation. \cite{Defferrard2016} used a simple scalar per vertex as graph signal, resulting in $X \in \mathbb{R}^N$. When using a feature vector for every vertex as described above, we can transfer the learned parameters $\theta_k$ into learnable parameter matrices $\Theta_k \in \mathbb{R}^{M \times F}$. Here, $M$ corresponds to the input dimension and $F$ to the output dimension of the feature representation, which is processed by Eq.~\ref{eq:cheb_layer}. For a more detailed description of the underlying theory, the reader is referred to the literature~\citep{Hammond2011,Defferrard2016,Bronstein2017}.

\subsubsection*{Implications of spectral graph convolution}
The previously defined approximations and conversions allow using a spectral convolution operation without the explicit and computationally expensive necessity to transfer the signal into Fourier space. However, the underlying assumption of the described process is an operation in the spectral domain. The learned representations are therefore sensitively depending on the eigenbasis defined by the full graph signal $X$. All data that should be part of the processing of the trained network, therefore, has to be part of the graph structure already during training time. A modification of the graph or inference on new unseen data would change the eigenbasis representation and an application of the trained network would become invalid. We can derive a few strong implications from this situation:
\begin{itemize}
\item Within the scope of semi-supervised training, the network shows high potential. Here, we assume that for a given amount of training data only a small ratio of samples has corresponding ground truth information $Y_{GT}$, e.g. a label for a vertex-wise classification task. Here, every vertex corresponds to a data sample, the vertices are connected by the edges $E$ of the graph. An example would be a data set of patients with a feature representation $X$, where every patient corresponds to a vertex in the graph and the edges between them are based on the similarity of additional patient metadata. For a standard supervised approach, the training data without label information would be difficult to use during training, for the graph convolution it however naturally becomes part of the training. It is assumed that the graph forms clusters of similar representations and groups together vertices with similar properties. Since the representations from neighboring vertices are aggregated into the feature representation $x_i$ of the node of interest, this clustering can stabilize and aid the prediction task, especially when the initial feature representation is less representative. Since the graph eigenbasis is incorporating the test data already, the trained network performs this task very effectively.
\item The necessity for all data to be present during training time also results in methodological drawbacks. Especially in the medical field, where it is required for a system to provide a diagnosis or treatment planning for new individual patients, the necessity of retraining of the full network for new data points is time-consuming and infeasible if e.g. the calculation power is not provided or the decisions are time-critical.
\item Even if all data is present during training time and the application does not require a frequent inference on new data points, depending on the number of considered vertices and the size of individual representations $x_i$ the network can run into memory problems quickly. Since the network needs to train on the full amount of data, a batch-wise training approach is not applicable which makes the processing of especially imaging or even volumetric data points nearly impossible for larger datasets.
\end{itemize}

\subsubsection*{Inductive Vertex Embedding}
\label{sec:theory_inductive}
To overcome the previously described limitations, we consider a spatial and therefore inductive approach. Based on the concepts of the methodology described in this section, also our proposed network structure will be based. The term inductive refers to a network type that is applicable for the inference of new unseen data samples that did not have to be part of the data set during training time. It, therefore, allows batch-wise training and is not limited to one graph structure. Since spatial approaches directly operate on the graph signal $X$, it is easier to assure the locality of the learned filters. Further, no transformation of the signal into Fourier space is necessary, no calculation of the graph Laplacian and its eigenbasis is required, which reduces complexity. In its core, for a node $v_i$ the presented methodology consists of an effective aggregation of representations $x_j$ from neighboring vertices $v_j$ into the representation of interest $x_i$.\\
Within the methodology GraphSAGE, \cite{Hamilton2017} provided a basic spatial graph network approach to perform the representation aggregation. The methodology is based on the direct usage of the weighted adjacency matrix $W$. Again, our graph is represented as a number of vertices $V$ and corresponding graph signal $X$ and connecting edges $E$ with weighted adjacency matrix $W$, forming the graph $G(V, E, W)$. Within this approach the edges are assumed to be undirected and binary, therefore $W$ consists only of values $0$ and $1$, where $w_{ij}=1$ represents the presence of an edge between the vertices $v_i$ and $v_j$.\\
The neighborhood aggregation is performed by a function $\Omega$ considering the direct (1-hop) neighborhood $N(i)$ of vertex $v_i$. For the $k^{th}$ layer of the network, we receive the following aggregation:
\begin{equation}
    x_{N(i)}^k = \Omega(\{ x_u^{k-1},\forall u \in N(v) \})
\end{equation}
Since the graph and data set can have a substantial size and low sparsity, it might not be feasible to aggregate the complete neighborhood $N(i)$. Therefore, only a constant sub-size $S$ of the neighborhood is considered. If the amount of $x_j \in N(i)$ is larger than $S$, a random subset of $N(i)$ is chosen, if the amount is smaller than $S$, vertices within $N(i)$ are randomly repeated until the necessary amount $S$ is reached. Like this, we can control the computational cost of the operation.\\
After receiving the aggregated neighborhood representation $x_{N(i)}^k$, it is concatenated with the current input representation $x_i^{k-1}$ of the vertex of interest $v_i$. Then, a learnable linear transformation $\Theta$ is applied to the full vector.
\begin{equation}
    x_i^k = \sigma (\Theta^k \cdot |x_i^{k-1}||x_{N(i)}^k|) ~,
\end{equation}
where $\sigma$ refers to a non-linear activation function and $|...||...|$ refers to the concatenation operation. This operation performs a learned and scalable aggregation of direct neighborhood (1-hop) representations into the individual representations of $X$. By consecutively applying $K$ GraphSAGE layers, we therefore effectively perform aggregation of the $K$-hop neighborhood. The learned filters are localized and trained on local graph signal representations directly, allowing a transfer on new unseen data points.

\subsubsection*{Attention-based feature aggregation} \label{sec:theory_gat}
Although the GraphSAGE network is able to learn a feature aggregation, the considered neighborhood is treated equally, ignorant to the potentially higher importance of certain representations to the update step. To solve this problem, \cite{Velickovic2018} proposed an evolution to this approach within the concept of Graph Attention Networks (GAT). Instead of directly aggregating the complete neighborhood $N(i)$ of a vertex $v_i$, an attention coefficient is calculated for every neighbor of $v_i$ individually using a shared linear transformation $\Theta \in \mathbb{R}^{M \times F}$, where $F$ is the dimension of the new representation. For a representation $x_i$, we calculate $\epsilon_{ij} = a(\Theta x_i, \Theta x_j)$, where $a$ is an attention function.~\cite{Velickovic2018} use a linear transformation $a^T \in \mathbb{R}^{2F}$ applied to the concatenated representations $\Theta x_i$ and $\Theta x_j$. Since the size of the resulting attention coefficient can vary largely depending on the input representations, a softmax is applied to all learned coefficients to receive the normalized attention coefficients $\alpha$:
\begin{equation}
    \alpha_{ij} = \frac{\exp(\sigma(a^T|\Theta x_i||\Theta x_j|))}{\sum_{k \in N(i)} \exp(\sigma(a^T|\Theta x_i||\Theta x_k|))}
\end{equation}
Using these coefficients, a propagation rule for the GAT layer can be defined. Here, the attention coefficients $\alpha_{ij}$ are used to weight the different contributions $x_j$ for the update of $x_i$ based on the learned attention corresponding to their importance. To statistically stabilize the prediction, multiple attention heads are used:
\begin{equation}
    x_i' = ||_{p=1}^P \sigma(\sum_{j \in N(i)} \alpha_{ij}^p \Theta^p x_j) ~,
\end{equation}
where $P$ refers to the number of used heads. The representations resulting from the $P$ heads are concatenated ($||$) to receive $x_i' \in \mathbb{R}^{P \cdot F}$ and transferred to the next GAT layer. Only the last layer performs an averaging of the different received representations. For a more detailed description of the GAT network, we refer to the literature \citep{Velickovic2018}.



\normalsize

\bibliography{references}


\end{document}